\documentclass[letterpaper]{article} 
\usepackage{aaai2026}  
\usepackage{times}  
\usepackage{helvet}  
\usepackage{courier}  
\usepackage[hyphens]{url}  
\usepackage{graphicx} 
\usepackage{natbib}  
\usepackage{caption} 
\usepackage{algorithm}
\usepackage{algorithmic}

\usepackage{newfloat}
\usepackage{listings}
\usepackage{xcolor}
\usepackage{colortbl}
\usepackage{booktabs}
\usepackage{multirow}

\usepackage{tocbibind}  
\usepackage{titlesec}   
\usepackage{minitoc}  
\usepackage{tocbibind}
\usepackage{amsmath}

\DeclareCaptionStyle{ruled}{labelfont=normalfont,labelsep=colon,strut=off} 
\floatstyle{ruled}
\newfloat{listing}{tb}{lst}{}
\floatname{listing}{Listing}

\providecommand{\eg}{\textit{e.g.}\@}
\providecommand{\etc}{\textit{etc.}\@}

\title{Res-Bench: Benchmarking the Robustness of Multimodal Large Language Models to Dynamic Resolution Input}
\author{
    Chenxu Li\equalcontrib \textsuperscript{\rm 1},
    Zhicai Wang\equalcontrib \textsuperscript{\rm 1}, 
    Yuan Sheng\textsuperscript{\rm 1},
    Xingyu Zhu\textsuperscript{\rm 1},
    Yanbin Hao\textsuperscript{\rm 2}\thanks{Corresponding author},
    Xiang Wang\textsuperscript{\rm 1},
}
\affiliations{
    \textsuperscript{\rm 1} University of Science and Technology of China \\
    \textsuperscript{\rm 2} Hefei University of Technology\\
    \{cxli2024,wangzhic,yuansheng,xyzhuxyz\}@mail.ustc.edu.cn,haoyanbin@hotmail.com,xiangwang1223@gmail.com

}

\usepackage{bibentry}

\begin{document}

\maketitle

\begin{figure*}
  \includegraphics[width=\textwidth]{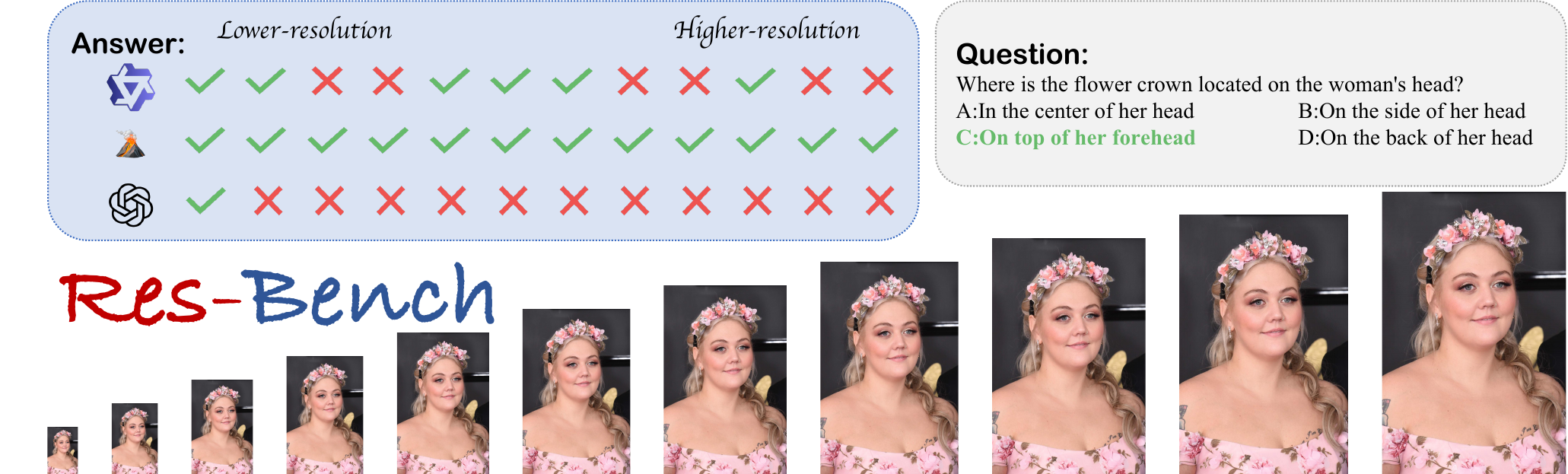}
      \caption {\textbf{An Example of ResBench.} There is a lack of resolution robustness in MLLMs, where models' performance on the given task fluctuates unpredictably as the input resolution increases.}
  \label{figure1}
\end{figure*}

\begin{abstract}
Multimodal Large Language Models (MLLMs) increasingly support dynamic image resolutions. However, current evaluation paradigms primarily assess semantic performance, overlooking the critical question of resolution robustness - whether performance remains stable across varying input resolutions. To address this gap, we introduce \textbf{Res-Bench}, a comprehensive benchmark comprising 14,400 samples across 12 resolution levels and six core capability dimensions. We designed a novel evaluation framework that goes beyond traditional accuracy metrics to capture performance stability. This framework introduces multiple robustness metrics: Spearman's correlation for assessing resolution-performance trends, and Absolute/Relative Continuous Error (ACE/RCE) for measuring performance volatility. Using these metrics, we conducted a large-scale evaluation of leading MLLMs. Our analysis encompasses: (1) model-centric and task-centric robustness examination, (2) investigation of preprocessing strategies including padding and super-resolution, and (3) exploration of fine-tuning for stability enhancement. Our code is available at \url{https://github.com/cxli2333/Res-Bench}.


\end{abstract}

\section{Introduction}

Visual perception and reasoning constitute fundamental components of human perceptual and cognitive capabilities. Notably, humans demonstrate remarkable robustness in processing and interpreting visual information across varying image resolutions. 
This is largely attributable to the adaptive capabilities of our visual system and the brain's cognitive completion mechanisms \cite{vision_book, mental_imagery}. Specifically, we can adaptively focus on important regions of an image, and for low-resolution images, our brains can fill in missing details based on past experiences and prior knowledge \cite{walsh1998perceptual}. The quest to equip machines with such sophisticated perceptual abilities is a central theme in modern AI research. However, while multimodal large language models have demonstrated exceptional capabilities across various domains, it remains unclear how well MLLMs can achieve resolution robustness comparable to human visual systems. 
%

Contemporary advanced MLLMs have incorporated sophisticated designed architectures to enable support for dynamic resolution input. These strategies generally fall into two categories: (1) \textbf{Native dynamic processing method}, such as the Vision Transformer (ViT) \cite{vit} in Qwen2.5-VL \cite{qwen2vl} which employs Multimodal Rotary Position Embedding (MRoPE) and window attention to process images at their native resolution; and (2) \textbf{Patch-based method}, utilized by models like InternVL2.5 \cite{InternVL2.5}, LLaVA-OneVision \cite{llava-ov} and LLaVA-UHD \cite{guo2024llava-uhd}, which segment high-resolution images into smaller sub-images and thumbnails for individual processing.
Current research paradigms predominantly focus on benchmarking model performance at the semantic level, including visual perception \cite{blink}, visual question answering (VQA) \cite{vqa}, visual reasoning \cite{mathvista}, \etc. However, insufficient attention has been directed to the robustness discussion at the visual processing level.
For each MLLM, three critical ingredients contribute to the resolution robustness:  (1) the processing mechanism for dynamic resolution input entailed in the vision encoder \cite{swin_transformer},  (2) the level of question spanning both perceptual and cognitive dimensions \cite{perception_reasoning}, and (3) the distribution of training data \cite{survey_mllm_data}.

To systematically investigate this gap, we introduce Res-Bench, a new benchmark specifically designed to evaluate the resolution robustness of MLLMs. Res-Bench is constructed based on 1,200 high-quality image-question pairs carefully selected by humans. For each image, we generate a series of lower-resolution versions via downsampling. The benchmark features a diverse set of questions across six primary categories and 15 fine-grained subcategories (see Section 3.2), covering a wide range of visual-linguistic tasks. We employ a suite of four metrics to comprehensively assess model behavior: overall performance is measured by accuracy, the performance trend by Spearman's correlation coefficient, and stability by Absolute/Relative Continuous Error (ACE/RCE), as detailed in Section 3.3. 

We conduct extensive experiments on Res-Bench to evaluate their robustness to visual inputs. Additionally, for the obtained low-resolution images, we employ data preprocessing methods like white-padding and super-resolution, aiming to validate their effectiveness in both enhancing absolute performance and improving resolution robustness\cite{DBLP:conf/aaai/ZhuWLHL024, DBLP:conf/nips/ZhuZ00HZ24}. The evaluation covers multiple proprietary models (GPT-4o( \cite{gpt4o}, and Gemini-1.5 Pro \cite{gemini}) and open-source models (\eg LLaVA-OneVision \cite{llava-ov}, Qwen2.5-VL \cite{qwen2vl}, mPLUG-Owl3 \cite{mplug}, InternVL2.5\cite{InternVL2.5}, \etc). We find there is a pervasive lack of resolution robustness. To understand the underlying factors contributing to this instability, we conducted an in-depth analysis of the experimental results, revealing several novel findings that pose new challenges for the development of MLLMs:

\begin{itemize}

\item \textbf{Architectural Trade-offs:} Native processing methods tend to achieve higher peak performance but are less robust to resolution changes. Conversely, patch-based methods demonstrate better robustness but at the cost of lower overall performance.

\item \textbf{Task-Dependent Robustness:} Our analysis reveals that resolution robustness is highly task-dependent. While some tasks are largely immune to resolution degradation, others show a strong, positive correlation between input quality and performance.

\item \textbf{Enhanced Visual Evidence via Preprocessing:} Information-free padding shows that performance peaks at a moderate input size, declining if the padding becomes excessive. Super-resolution provides a more significant boost by restoring visual information, outperforming padding.

\item \textbf{Enhancing Robustness via Fine-tuning:} Fine-tuning on a resolution-balanced dataset can notably enhance a model's resolution robustness. The enhanced resolution robustness of the fine-tuned model shows generalization to out-of-distribution data. 
\end{itemize}

\section{Related Work}

\begin{table*}[t]

\centering

    \begin{tabular}{cccc}
    \toprule
    Benchmark&Dimension&Task&Robustness\\
    \midrule
    MLLM-COMPBENCH&8&Multi-images MCQ& Relative\\
    MMVU&12&Multi-images MCQ&Relative\\
    RobustBench&15&Single-image MCQ &Absolute\\
    NaturalBench&1&Single-image MCQ &Relative\\\midrule
    
    Res-Bench (ours) &15&Single-image \& Multi-images MCQ, VQA&Absolute, Relative\\
    \bottomrule

    \end{tabular}
    \caption{\textbf{Robustness Benchmarks and Res-Bench.} The comparison highlights key differences in scope, including task diversity and the robustness evaluation methodology.}
    \label{table1}
\end{table*}

\textbf{Multimodal Large Language Model}.
A fundamental challenge in Multimodal Large Language Models (MLLMs) \cite{qwen2vl, llava-ov, gemini} lies in effectively processing visual information, particularly at high resolutions where fine-grained details are critical.
Early models addressed this by resizing images to a low, fixed resolution, a method prone to distorting critical details necessary for tasks like OCR or fine-grained perception \cite{clip, qwenvl}.
To address these issues, several common solutions have emerged:
(1) Native resolution processing methods, such as those used in Qwen2.5-VL \cite{qwen2vl} and Kimi-VL \cite{kimi}.
(2) Patch-based method: employed by models like LLaVA-OneVision \cite{llava-ov}, InternVL2.5 \cite{InternVL2.5}, and DeepSeek-VL2 \cite{deepseekvl}.
While both approaches handle high-resolution inputs, the trade-offs between them regarding resolution robustness remain largely unexplored.  This paper systematically investigates the impact of these distinct architectural choices on model stability\cite{DBLP:journals/corr/abs-2507-03657, DBLP:conf/mm/ZhuZTWHZ24}.

\noindent\textbf{MLLM Benchmarks}.
The evaluation of MLLMs has rapidly evolved, moving from single-task benchmarks like VQA \cite{vqa, vizwiz}, caption \cite{coco, flickr30k} to comprehensive, multi-faceted evaluation suites. Recent efforts have produced a rich ecosystem of benchmarks targeting a wide array of high-level capabilities, including:
Visual perception \cite{blink},
Chart reasoning \cite{chartqa, chartqapro, charxiv},
Mathematical reasoning \cite{mathvista,mathverse},
Spatial reasoning \cite{spatialeval, mvot, bsa},
Expert-level multimodal understanding \cite{mmmu, mmmupro},
Optical Character Recognition (OCR) capabilities \cite{OCRbench, ccocr}.
Additionally, comprehensive benchmarks \cite{seed, mmstar} have been developed to evaluate overall multimodal intelligence. While their source images may have varying native resolutions, these benchmarks ignore the impact of the resolution distribution.

\noindent\textbf{Robustness Evaluation}.
Evaluating the robustness of MLLMs—their ability to maintain stable performance against input perturbations—is a critical area of research.
Existing work has primarily focused on several key dimensions: (1) robustness against semantic-level perturbations \cite{MAD-Bench, MMVU}; (2) robustness against stylistic or natural variations in images \cite{benchlmm, NaturalBench}; and (3) robustness against common image corruptions \cite{robustbench, rbench, zhu2025enhancing}.
A direct comparison of these approaches, summarized in Table \ref{table1}, reveals a shared limitation: while they comprehensively examine model stability, they primarily test against content-level corruptions and adversarial inputs. The impact of fundamental visual properties like resolution on model performance remains a critical, under-explored area.
We introduce \textbf{Res-Bench} to systematically fill this gap, providing the first dedicated benchmark for evaluating MLLM resolution robustness.

\begin{figure*}[t]
	\centering
	\includegraphics[width=\linewidth]{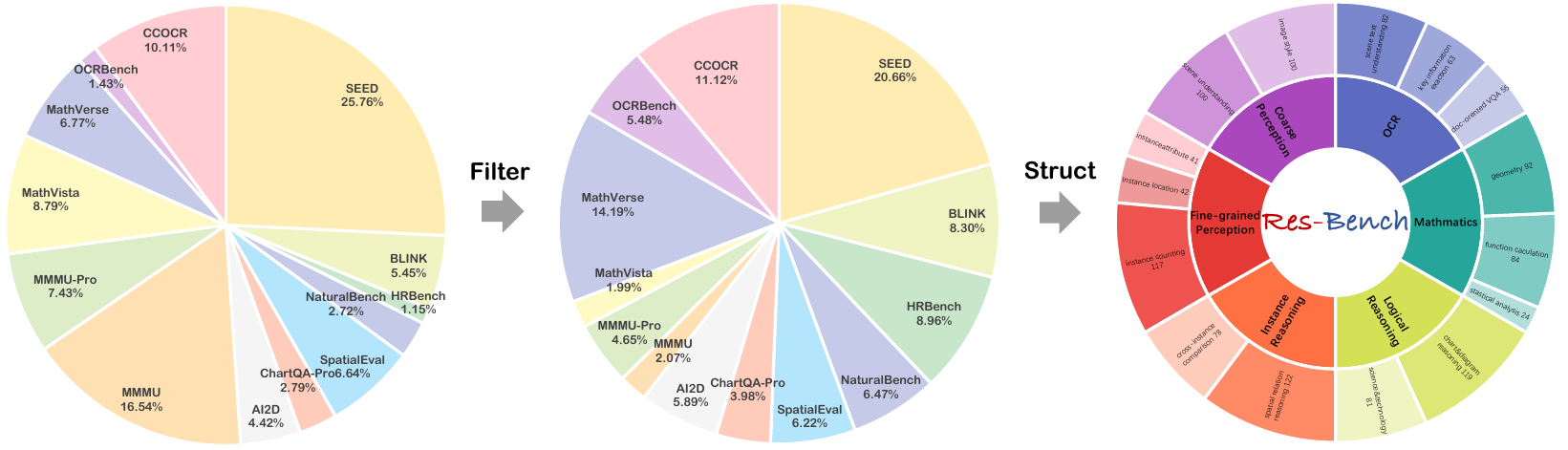}
	\caption{\textbf{ResBench Dataset Construction Overview.} The pie charts illustrate the proportional contribution of each source dataset before and after our filtering process. The final sunburst chart shows the composition of the selected data, organized by our six core capability dimensions and 15 sub-tasks.}
	\label{figure2}
\end{figure*}


\section{Res-Bench}\label{section3}
We present our dataset, Res-Bench, a comprehensive benchmark for evaluating the dynamic resolution robustness of MLLMs. This section details the construction of Res-Bench, beginning with our multi-stage data collection process in Section 3.1. In Section 3.2 we provides an overview of the dataset's composition and statistical properties. In Section 3.3, we introduce four benchmark-specific metrics designed to assess model robustness.

\subsection{Data Collection Process}\label{section3.1}
Res-Bench is constructed through a rigorous, multi-stage pipeline designed to ensure data quality, diversity, and relevance. The process begins with establishing strict collection standards, followed by stringent automated filtering and expert manual verification.

\noindent\textbf{Data Collection Standards.} We defined three core principles for sample selection: 
\textbf{(1) Visual Dependency.}
When provided with only text input, the model fails to correctly answer the sample questions.
\textbf{(2) Diversity.} The dataset must span a wide range of MLLM capabilities—from low-level perception (\eg, text recognition, counting) to high-level reasoning (\eg, spatial relationships, mathematics)—to enable a holistic evaluation.
\textbf{(3) Pristine Image Quality.} Source images must be of high resolution and quality to serve as a reliable, artifact-free baseline. This is essential for systematically studying the effects of controlled resolution degradation via downsampling.

\noindent\textbf{Preliminary Data Filtering.} 
To establish a comprehensive and high-quality benchmark, we selected 13 existing benchmarks as data sources. These include benchmarks designed for general evaluation of MLLMs (\eg, SEED \cite{seed}) as well as specialized benchmarks for assessing specific capabilities of MLLMs (including AI2D \cite{ai2d}, HR-Bench \cite{hrbench}, NaturalBench \cite{NaturalBench}, SpatialEval \cite{spatialeval}, MMMU \cite{mmmu}, MMMU-Pro \cite{mmmupro}, ChartQA-Pro \cite{chartqapro}, BLINK \cite{blink},  MathVerse \cite{mathverse}, MathVista \cite{mathvista}, OCRBench \cite{OCRbench}, CCOCR \cite{ccocr}). To ensure the data consists exclusively of high-quality images, we first conducted preliminary filtering by selecting all high-resolution images (with both width and height exceeding 1344 px) as the foundational dataset. Building upon this foundation, we further refined the dataset by leveraging powerful models to select samples that meet visual dependency criteria. Specifically, we employed GPT-4o \cite{gpt4o} and InternVL-2.5-72B \cite{InternVL2.5} as evaluators for this filtering process. We conducted text-only input evaluations by providing textual questions without any accompanying images, while employing circular evaluation to mitigate random biases. A sample was automatically discarded if both models could answer it correctly without the image, as this indicated weak visual dependency. After this filtering process, the number of candidate samples was reduced from 69840 to 2868.

\noindent\textbf{Manual Verification.}
To further ensure data quality, we performed human expert review on the pre-filtered samples. The evaluation criteria remained consistent with the aforementioned standards. Additionally, we further optimized the data distribution to ensure better balance. To comprehensively evaluate the capabilities of large models across diverse question types and facilitate systematic assessment, we further refined the phrasing of questions for a subset of the data. For instance, we restructured the single-image multi-question format in NaturalBench \cite{NaturalBench} to a single-question multi-image paradigm, while also refining certain question formulations in CCOCR \cite{ccocr}. We then systematically generated the full Res-Bench dataset by creating multiple resolution versions for each image, resulting in a total of 14,400 data instances for evaluation.
Our data filtering pipeline is illustrated in Figure \ref{figure2}.

\subsection{Comprehensive Dataset Analysis}\label{section3.2}

Our benchmark comprehensively evaluates 6 core capability dimensions with 15 fine-grained sub-capabilities, comprising 200 high-quality samples per dimension. The data encompasses diverse question formats including:
Single-image multiple-choice questions (MCQ),
Multi-image MCQ,
VQA.
We systematically generated 14,400 images across 12 resolution groups  through controlled resizing of the 1,200 original samples, aligning each image to fixed target dimensions along its longer edge. Res-Bench includes six core capabilities (including Coarse Perception, Fine-grained Perception, Instance Reasoning, Logical Reasoning, Mathematical, OCR) and 15 sub-capabilities. The distribution across capability dimensions is as shown in Figure \ref{figure2}. More detailed capability definitions and cases can be seen in the appendix.

\subsection{Evaluation Metrics}\label{section3.3}

\begin{table*}[t]

  \centering
  \small  
  \setlength{\tabcolsep}{1mm}
  {
  \begin{tabular}{lcccccccccccccccc}
    \toprule
        \multirow{2}{*}{Model}&\multicolumn{12}{c}{Resolution of Input}
        &\multirow{2}{*}{$Acc_{avg.}\uparrow$}
        &\multirow{2}{*}{$\rho$ $\uparrow$}
        &\multirow{2}{*}{ACE$\downarrow$}
        &\multirow{2}{*}{RCE$\downarrow$}\\
        &112&224&336&448&560&672&784&896&1008&1120&1232&1344&\\
    \midrule
    
    \rowcolor{gray!30}\multicolumn{17}{c}{Proprietary Model}\\
    Gemini1.5-Pro&0.420&0.482&\textbf{0.517}&\textbf{0.571}&\textbf{0.584}&\textbf{0.590}&\textbf{0.592}&0.596&0.596&\textbf{0.591}&0.603&0.595&\textbf{0.561}&0.895&0.201&0.358\\
    GPT-4o&\textbf{0.423}&0.474&0.498&0.510&0.522&0.534&0.548&0.548&0.564&0.557&0.560&0.567&0.526&0.972&0.157&0.299\\
    
    \rowcolor{gray!30}\multicolumn{17}{c}{Open-source Model}\\
    Qwen2.5-VL\textsuperscript{\dag}&0.362&0.445&0.504&0.532&0.565&0.575&0.582&\textbf{0.598}&\textbf{0.607}&0.586&\textbf{0.609}&\textbf{0.599}&0.547&0.951&0.299&0.547\\
    Kimi-VL\textsuperscript{\dag}&0.369&0.415&0.469&0.512&0.541&0.542&0.547&0.568&0.559&0.567&0.565&0.573&0.519&0.951&0.226&0.437\\
    LLaVA-OV\textsuperscript{\ddag}&0.420&\textbf{0.489}&0.505&0.515&0.529&0.525&0.532&0.537&0.541&0.555&0.542&0.538&0.519&0.944&0.159&0.306\\
    InternVL-2.5\textsuperscript{\ddag}&0.403&0.460&0.498&0.533&0.546&0.552&0.553&0.557&0.551&0.559&0.559&0.556&0.527&0.909&0.170&0.323\\
    MiniCPM-o-2.6\textsuperscript{\ddag}&0.363&0.436&0.472&0.503&0.534&0.539&0.540&0.551&0.563&0.572&0.572&0.585&0.510&\textbf{1.000}&0.222&0.428\\
    mPLUG-Owl3&0.333&0.355&0.379&0.380&0.369&0.391&0.396&0.400&0.410&0.405&0.407&0.404&0.385&0.916&\textbf{0.109}&\textbf{0.282}\\
    \bottomrule
  \end{tabular}
  \caption{\textbf{Main Results.} The central columns show model accuracy at each of the 12 specific resolutions. The final four columns present the overall average accuracy and three robustness metrics. \textsuperscript{\dag} means the models directly process images at their native dynamic resolution. \textsuperscript{\ddag} means the models process high-resolution images by dividing them into smaller patches or sub-images. The best  results are highlighted in \textbf{bold}, respectively. $\uparrow$: higher is better. $\downarrow$: lower is better.}
  \label{table2}
}
  
\end{table*}

To quantify model robustness against resolution changes, we employ a suite of four complementary metrics.

\noindent\textbf{Accuracy.} We measure the model's fundamental capability on a given task at a specific resolution level. The accuracy at a resolution $res$ , denoted as$Acc_{res}$ is calculated as:
\begin{equation}
    Acc_{res} = Score(GT,MLLM(I_{res})),
\end{equation}
where $M(I_{res})$ is the model's prediction for an image $I$ at resolution $res$
$G$ is the ground truth, and the $Score$ function returns 1 for a correct answer and 0 otherwise. To gauge the model's overall performance across all tested resolutions, we then compute the \textbf{Average Accuracy}:
\begin{equation}
    Acc_{avg} = \frac {1}{N_{res}} \sum_{i=1}^{N_{res}} Acc_{res_i},
\end{equation}
where $N_{res}$ is the total number of resolution levels (12 in Res-Bench), and $res_i$ denotes the $i$-th resolution.

\noindent\textbf{Spearman's Correlation Coefficient.} 
While $Acc_{avg}$ provides an overall score, it does not capture the relationship between resolution and performance. The Spearman's correlation coefficient assesses the ordinal relationship between two variables. We treat the rank of the resolution and the rank of the accuracy as the two variables to assess their correlation. The detailed definition is as follows:
\begin{equation}
    \rho = 1 - \frac{6 \sum d_i^2}{n(n^2 - 1)},
\end{equation}
where $d_i$ denotes the rank difference, and $n$ denotes the resolution-sample size, which is 12 in our Res-Bench.

\noindent\textbf{Absolute Continuous Errors.} When two models may have equal average accuracy and Spearman's correlation coefficient but differ in the fluctuation magnitude of their accuracy as resolution changes, it is necessary to quantify such fluctuations. We define Absolute Continuous Errors(ACE) as the sum of absolute differences in accuracy between adjacent resolutions:
\begin{equation}
    ACE = \sum_{i=1}^{n-1}\left|Acc_{res_{i+1}}-Acc_{res_i}\right|,
\end{equation}
where $res_{i}$ means the $i$-th resolution.

\noindent\textbf{Relative Continuous Errors.} To enable fair comparison between models with different capability levels, RCE normalizes the fluctuation by the average accuracy:
\begin{equation}
    RCE = \frac{ACE}{Acc_{avg}}.
\end{equation}
RCE allows for a fairer comparison of robustness between models with different overall capability levels.
\section{Experiment}\label{section4}

In this section, we conduct a comprehensive experimental evaluation to assess the resolution robustness of state-of-the-art MLLMs using our proposed Res-Bench benchmark and metrics. We begin by detailing our experimental setup. Section 4.1 presents an analysis of the results from eight leading models on Res-Bench, providing a deeper, multi-faceted analysis of these results. Section 4.2 investigates the impact of data preprocessing techniques. Section 4.3 explores the potential of fine-tuning as a strategy to enhance robustness. Detailed evaluation criteria can be seen in Appendix.

\noindent\textbf{Baseline.} We use Res-Bench to evaluate several leading MLLMs. This includes two powerful proprietary models, GPT-4o \cite{gpt4o} and Gemini 1.5-Pro \cite{gemini}, and six powerful open-source models employing different dynamic resolution strategies. Qwen2.5-VL \cite{qwen2vl} , Kimi-VL-Instruct \cite{kimi} directly process native high-resolution images, while InternVL-2.5 \cite{InternVL2.5}, LLaVA-OneVision \cite{llava-ov} , MiniCPM-o-2.6 \cite{minicpm} use patch-based method. We also tested model that do not support dynamic resolution for single images, mPLUG-Owl3 \cite{mplug}.

\subsection{Experiment Analysis}\label{section4.1}

\begin{figure*}[t]
	\centering
        \includegraphics[width=\linewidth]
	{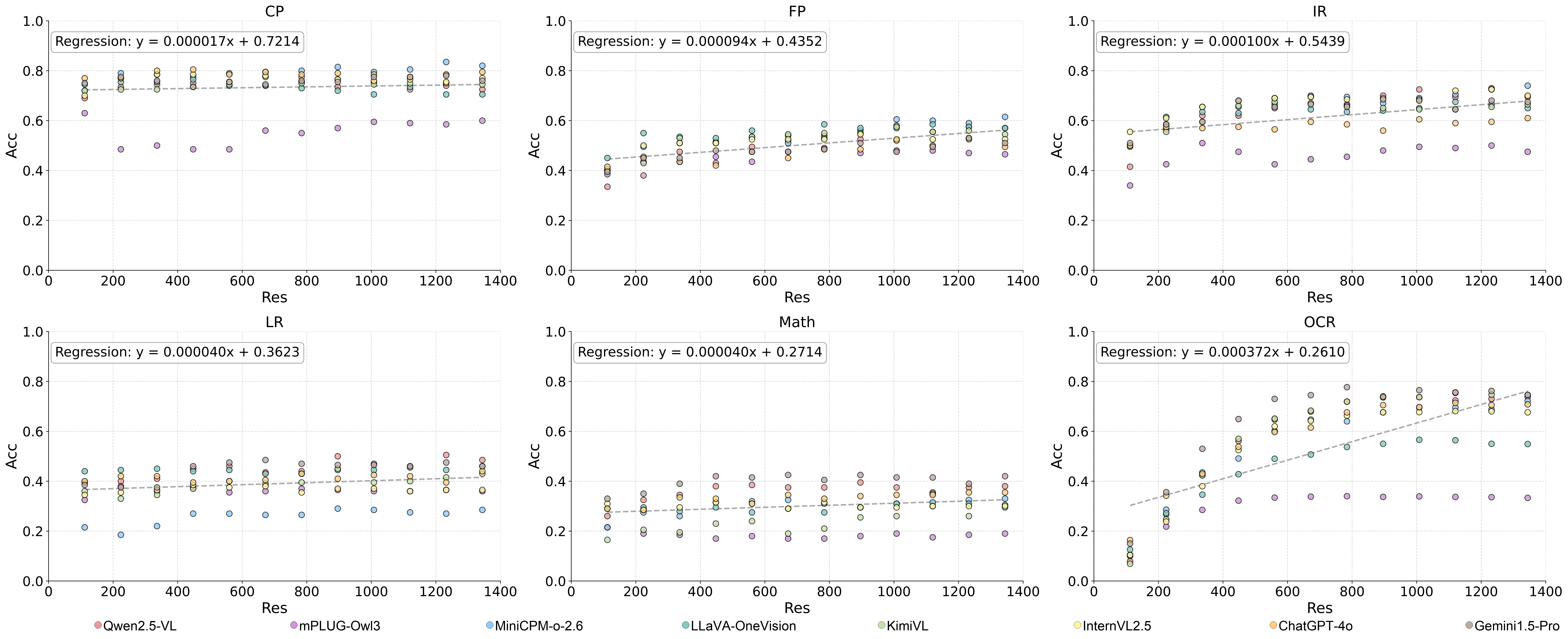}
	\caption{\textbf{Task-level Results.} We show the result with different input resolution across six core tasks. To better show the difference of resolution robustness among tasks, we show the results of linear regression on all data points in each subplot. CP: Coarse-grained Perception, FP: Fine-grained Perception, IR: Instance Reasoning, LR: Logical Reasoning, Math: Mathematical Reasoning, OCR: Optical Character Recognition.}
    \label{figure3}
\end{figure*}

\textbf{Main Result Analysis.}\label{section4.1.1}
The overall performance of the evaluated models on Res-Bench is presented in Table \ref{table2}. Our findings indicate that Res-Bench poses a significant challenge to current state-of-the-art MLLMs. The top-performing models, Gemini-1.5 Pro and Qwen2.5-VL, achieve an average accuracy($Acc_{avg}$) below 60\%. A central finding is the pervasive lack of resolution robustness. Most models exhibit unstable performance across different resolutions. This is highlighted by the metrics for Qwen2.5-VL, which, despite its high accuracy, shows the most fluctuation with the worst ACE and RCE scores among all models. In stark contrast, MiniCPM-o-2.6 demonstrates a perfectly monotonic performance trend ($\rho=1$), where accuracy strictly increases with resolution, though its absolute accuracy is not the highest. This reveals a clear and critical trade-off between peak performance and robustness.

\noindent\textbf{Model-centric Analysis.}
We now dissect the results from a model-centric perspective, focusing on how architectural choices influence resolution robustness:
proprietary models: A distinct trade-off is also visible between the two leading proprietary models. Gemini-1.5 Pro achieves one of the highest accuracy scores but demonstrates moderate robustness. Conversely, GPT-4o, while having a slightly lower average accuracy, exhibits superior stability with more favorable robustness metrics (lower ACE/RCE).
The open-source models reveal a clear divergence based on their high-resolution processing strategy:
\textbf{(1) Native dynamic processing models:} These models, which process images at their native resolution, generally achieve higher peak accuracy, particularly on high-resolution inputs. However, they are highly sensitive to resolution degradation, resulting in poor performance at lower resolutions and, consequently, high fluctuation scores (ACE/RCE).
\textbf{(2) Patch-based processing models:} These models, which rely on image patches and thumbnails, demonstrate superior robustness. Their performance is more stable across resolutions, and they perform comparatively better on low-resolution inputs. The trade-off is a lower performance ceiling, as their accuracy does not scale as effectively with increasing resolution.
\textbf{(3) Fix-resolution models:} mPLUG-Owl3, which lacks a dynamic resolution mechanism, lags significantly in overall accuracy. While its ACE and RCE scores appear low, we posit this is likely an artifact of its compressed performance range rather than an indicator of true robustness.

\noindent\textbf{Task-centric Analysis.}
In Figure \ref{figure3}, analyzing performance by capability dimension reveals that sensitivity to resolution varies significantly across tasks, as clearly indicated by the linear regression trends for each task category.
For Coarse Perception (CP) tasks, the regression line demonstrates an almost flat slope ($y=0.000017x+0.7214$), indicating that these tasks are highly robust to resolution loss. Models maintain strong performance even at very low resolutions (e.g., 112px), and further increasing resolution yields negligible improvements. This suggests that holistic understanding requires minimal visual detail.
For Fine-grained Perception (FP), Instance Reasoning (IR), Logical Reasoning (LR), and Mathematical Reasoning (Math), the regression lines exhibit weak positive slopes . This indicates only a marginal correlation between resolution and accuracy ($\rho\approx0.5$). Furthermore, significant accuracy fluctuations between adjacent resolution steps are observed, highlighting the instability of model performance in these tasks. Merely increasing resolution does not guarantee better or more stable reasoning for these categories.
For OCR tasks, the regression slope is considerably steeper (
$y=0.000372x+0.2610$), reflecting a strong dependency on resolution. A distinct threshold effect is observed: performance degrades severely at lower resolutions, but once a minimal clarity threshold is reached, accuracy improves significantly with higher resolutions before plateauing. This suggests that OCR is highly resolution-sensitive but becomes robust once sufficient clarity is achieved. The linear regression trends across all tasks provide a key quantitative perspective on resolution robustness, reinforcing the task-dependent nature of model sensitivity to resolution changes.

\subsection{Analysis of Preprocessing Strategies}\label{section4.2}

\begin{figure}[htbp]
	\centering
        \includegraphics[width=\linewidth]{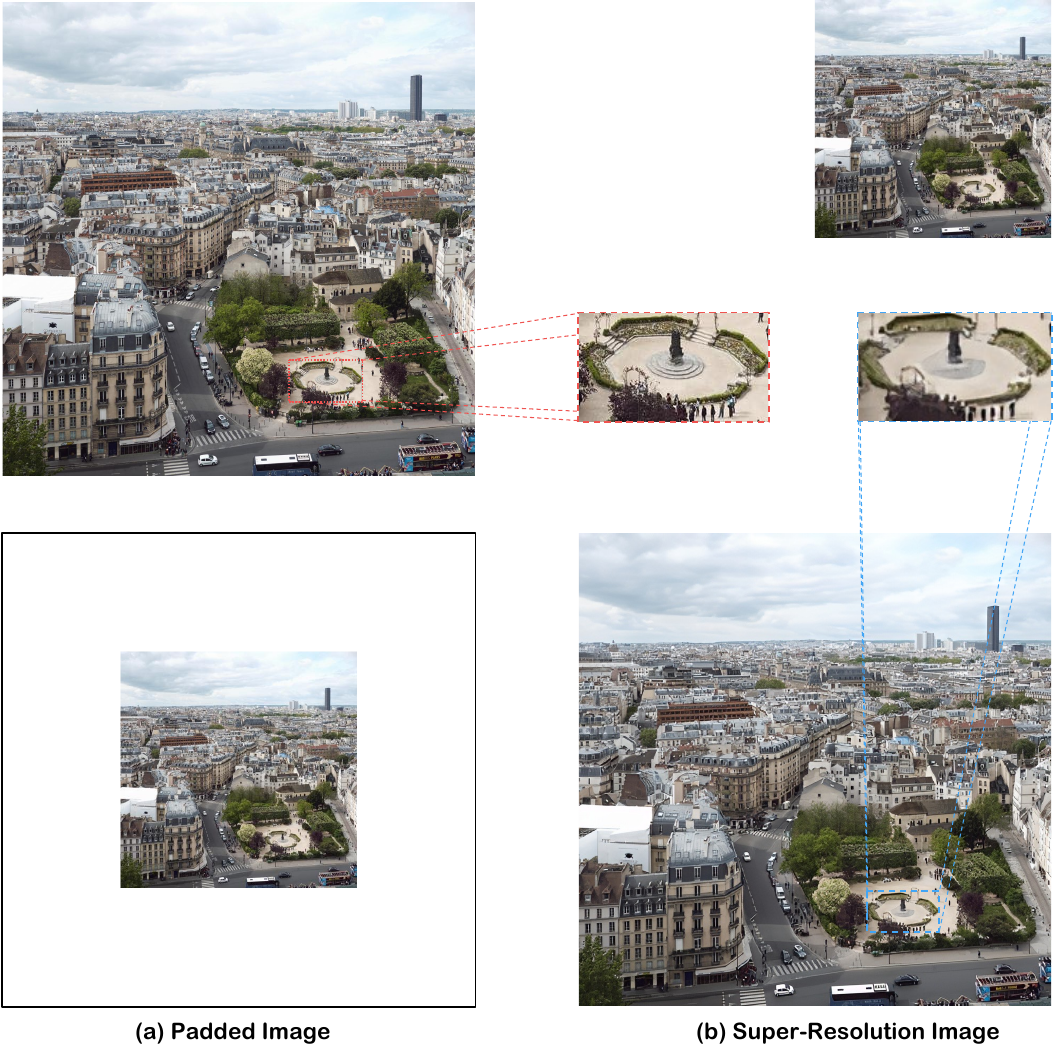}
	\caption{\textbf{Example of Preprocessing.} (a) Image padding by adding non-imformation pixels around images. (b) Image super-resolution using off-the-shelf SR models.}
        \label{figure4}
\end{figure}

To better understand the performance degradation on low-resolution images, we conducted experiments to disentangle two factors: reduced information content and shorter visual token length. We investigate this through two augmentation strategies: white-padding and super-resolution (SR). The example of processing strategies are shown in Figure \ref{figure4}.

\begin{table}[htbp]
\centering
\small  
\setlength{\tabcolsep}{1mm}
\resizebox{\linewidth}{!}{
\begin{tabular}{cllll}
\toprule
Strategy&\multicolumn{2}{c}{Padding}&\multicolumn{2}{c}{SR}\\
\cmidrule(lr){2-3} \cmidrule(lr){4-5}
Final Res&224&448&224&448\\
\midrule
    224&$0.445$&&$0.445$&\\
    336&$0.447_{(+0.002)}$&&&\\
    448&$0.448_{(+0.003)}$&$0.532$&$0.473_{(+0.028)}$&$0.532$\\
    672&$0.453_{(+0.008)}$&$0.550_{(+0.018)}$&\\
    784&$0.447_{(+0.002)}$&$0.544_{(+0.012)}$&&\\
    896&$0.438_{(-0.007)}$&$0.553_{(+0.021)}$&$0.478_{(+0.032)}$&$0.558_{(+0.026)}$\\
    1008&$0.443_{(-0.002)}$&$0.538_{(+0.006)}$&&\\
    1232&$0.434_{(-0.011)}$&$0.541_{(+0.009)}$&&\\
    1344&$0.440_{(-0.005)}$&$0.525_{(-0.007)}$&&\\
\bottomrule
\end{tabular}
}
\caption{\textbf{Impact of Image Padding and Super-Resolution.} Columns denote the initial image resolution (224px or 448px), while rows indicate the final resolution after processing. The baseline performance for each unprocessed initial image is shown on the diagonal where the initial and final resolutions are identical. SR:Super-Resolution.}
\label{table3}
\end{table}

\noindent\textbf{Isolating the Effect of Token Length via Padding.}
A high-resolution image, when processed by a ViT, typically results in a longer sequence of visual tokens than a low-resolution one. To isolate the effect of this token length, we designed an experiment where information content was held constant while token length was increased. We took low-resolution images (specifically, those at 224px and 448px) and padded them with zero-value pixels to match the dimensions of various higher resolution targets.
For both the 224px and 448px source images, padding to a moderately higher resolution leads to a performance improvement, even though no new visual information was added. However, this trend reverses when padding to the higher resolutions, where we observe a slight decrease in accuracy. It may be because of the higher visual/textual token length ratio, which increases the model's attention to visual input, making the model more reliant on visual input rather than relying on prior knowledge. We hypothesize that while a longer token sequence can be beneficial, an excessive number of non-information padded tokens can act as noise, diluting the model's attention on the actual visual content and impairing its performance.

\noindent\textbf{Restoring Information Content via Super-Resolution.}\label{section4.3}
We investigated whether algorithmically restoring visual details could recover lost performance more effectively. We employed DiffIR \cite{diffir}, a classical diffusion-based SR model, to upscale low-resolution images (224px and 448px) to higher resolutions (448px and 896px). Unlike padding, SR aims to intelligently ``fill in" missing information, enhancing image clarity. As shown in Table \ref{table3}, SR-enhanced images significantly outperform their original low-resolution counterparts and also surpass the performance of the simple padded images. 

\noindent\textbf{Discussion.} Our experiments reveal two key findings: (1) Token sequence length affects performance - moderate padding is beneficial but excessive padding is detrimental; (2) Visual information quality is crucial - while SR techniques improve performance significantly, they cannot fully match the quality of original high-resolution images. 

\subsection{Enhancing Robustness via Fine-tuning}


To explore whether resolution robustness can be learned, we investigated if fine-tuning on a mixed-resolution dataset could improve a model's stability. We fine-tuned the Qwen2.5-VL-3B model specifically on spatial reasoning tasks. The detailed experimental settings and hyperparameters are in the appendix.

\begin{table}[ht]
    \centering
    \begin{tabular}{ccccc}
    \toprule
    \multirow{2}{*}{Task}&\multicolumn{2}{c}{ACE$\downarrow$}&\multicolumn{2}{c}{$Acc\uparrow$}\\
    \cmidrule(lr){2-3} \cmidrule(lr){4-5}
    &w/o FT&FT&w/o FT&FT\\
    \midrule
    CP&$0.369$&$0.433_{+0.064}$&$0.678$&$0.670_{-0.008}$\\
    FP&$0.578$&$0.546_{-0.043}$&$0.510$&$0.495_{-0.015}$\\
    IR&$0.539$&$0.478_{-0.061}$&$0.454$&$0.607_{+0.153}$\\
    LR&$0.309$&$0.281_{-0.028}$&$0.469$&$0.445_{-0.024}$\\
    Math&$0.658$&$0.576_{-0.082}$&$0.266$&$0.295_{+0.029}$\\
    OCR&$1.230$&$1.281_{+0.051}$&$0.540$&$0.540_{+0.000}$\\
    \bottomrule
    \end{tabular}
    \caption{\textbf{The Impact of Fine-tuning.} The table compares the metrics of the model before (w/o FT) and after (FT) being fine-tuned on spatial reasoning tasks.}
    \label{table4}
\end{table}

\noindent\textbf{Results and Analysis.}
The results, detailed in Table \ref{table4}, show that this strategy was highly effective. The fine-tuned 3B model not only became more accurate but also significantly more robust in its target domain. On the `Instance Reasoning' dimension, the model's performance improved dramatically. Accuracy surged from 0.454 to 0.607, while the RCE dropped sharply from 0.539 to 0.478. This indicates a clear, simultaneous improvement in both capability and stability. The fine-tuning experiment also demonstrated the model's generalization ability. Overall average accuracy increased from 0.486 to 0.508, while both ACE (0.224 to 0.220) and RCE (0.461 to 0.433) decreased, confirming a general enhancement of robustness. There is also a Robustness-Performance Trade-off. Interestingly, on some related but out-of-domain tasks like `Fine-grained Perception', we observed a slight decrease in absolute accuracy. However, even in this case, the model's robustness improved, as evidenced by a lower RCE. This suggests that the model learned a more stable general representation, even if it came at the cost of peak performance on certain tasks it was not explicitly trained on.
this experiment demonstrates that mixed-resolution fine-tuning is a promising and effective strategy for directly enhancing the resolution robustness of MLLMs. It teaches the model to handle visual inputs more consistently across quality levels.

\section{Limitations and Future Work}
While this work provides the first systematic evaluation of MLLM resolution robustness, we acknowledge several limitations that open avenues for future research:

\noindent\textbf{Limited Range of Resolution Levels.} Our current Res-Bench evaluates resolutions up to approximately 1.5K. It does not include extremely high-resolution cases (e.g., 4K, 8K), which are becoming increasingly common. Evaluating models on such ultra-high-resolution inputs could reveal new challenges related to computational efficiency, architectural bottlenecks, and the ability to process an even greater density of information.

\noindent\textbf{Narrow Exploration of Preprocessing Techniques.} Our investigation into performance recovery was limited to a single diffusion-based super-resolution method (DiffIR). Future work could conduct a comparative study of a wider array of SR techniques, including GAN-based or transformer-based models. This could uncover which algorithmic approaches are most effective at restoring critical information for MLLMs and potentially lead to tailored SR models designed specifically for downstream multimodal tasks. 
\section{Conclusion}

In this paper, we introduce Res-Bench, a new benchmark designed to evaluate the critical yet overlooked dimension of MLLM resolution robustness. Constructed from 1,200 curated samples across 6 core capabilities and 12 resolution levels, Res-Bench is paired with novel metrics like ACE/RCE to quantify stability beyond simple accuracy. Our comprehensive evaluation reveals a pervasive lack of stability in state-of-the-art models. Our analysis identified key factors influencing this behavior, including a clear architectural trade-off between performance and robustness, and the impact of input token length. Crucially, we also demonstrated that robustness can be explicitly learned through mixed-resolution fine-tuning, offering a direct path for improvement. We believe Res-Bench provides a vital tool for the community, pushing the development of MLLMs towards the more reliable and adaptive intelligence characteristic of human perception.

\newpage

\bibliography{sections/main}
\newpage




\appendix

\section{Appendix}


\subsection{Discussions}

\subsubsection{Social Impacts}

Our work identifies the specific resolution thresholds at which MLLMs begin to fail. A potential negative impact is that malicious actors could exploit these identified failure points to craft low-resolution content specifically designed to evade automated safety and content moderation systems. To counteract this, we urge developers to use Res-Bench not only as a performance benchmark but also as a red-teaming tool to proactively identify and harden their models against these vulnerabilities.

\subsubsection{Ethical Considerations}

We have taken several steps during the construction of Res-Bench to mitigate potential ethical concerns.
First, the source data for Res-Bench is aggregated from 13 established and publicly available academic benchmarks (see Section 3.1). These source datasets are widely used in the research community and are generally presumed to be free of harmful, unethical, or offensive content.
Additionally, during our multi-stage filtering and manual verification process, our human experts were explicitly instructed to screen for and discard any samples that contained personally identifiable information (PII), offensive material, or other sensitive content. The question refinement process also aimed to keep all prompts neutral and strictly task-focused.
Finally, the core objective of Res-Bench is to evaluate the perceptual robustness of MLLMs against resolution degradation. The benchmark does not contain tasks that involve comparing or generating judgments on sensitive, ethical, or harmful subjects.

\subsubsection{License of Assets}

All thirteen datasets are publicly available. We release our Res-Bench under a Creative Commons Attribution 4.0 License (CC BY 4.0) to enhance global accessibility and foster innovation and collaboration in research.

\subsection{Experiment Details}

\subsubsection{Evaluation Criteria}
To ensure fair and consistent evaluation across all models, we established a rigorous evaluation protocol covering answer adjudication, reproducibility, and implementation details.
For MCQ, The ground truth is a single option letter (e.g., ``A''). We accept answers even if they are embedded in common output patterns like ``A: [content]'' or ``The correct answer is A.''. 
For mathematical problems, The ground truth is the final numerical result. We perform a direct substring search for the ground truth number within the model's entire generated output. This approach correctly credits answers that are part of a step-by-step reasoning process. 
For OCR tasks, The ground truth often consists of multiple words or a sentence. We tokenize both the ground truth and the model's output and evaluate performance based on word-level exact matching. 
For non-deterministic decoding, the results are averaged over three independent runs for open-source models. 
For proprietary models, due to their limited availability and increased compute time and cost, we only perform one run. 
Moreover, all experiments in this study are conducted within the same codebase modified from VLMEvalKit, utilizing NVIDIA A100 GPUs for non-API-based evaluation.

\subsubsection{Fine-tuning Experiment Setup}
We curated a training set of 628 high-quality images related to spatial reasoning from the SpatialEval and SEED datasets, ensuring no overlap with our Res-Bench test set. Crucially, to construct the full training dataset, we systematically processed each of the 628 source images, creating a version for all 12 resolution levels defined in Res-Bench. This process resulted in a comprehensive 7536-instance training set, with each image systematically presented across our 12 defined resolution levels to ensure robust learning. We then performed full-parameter fine-tuning on Qwen2.5-VL-3B with this dataset.

We conducted full-parameter fine-tuning on the Qwen2.5-VL-3B-Instruct model using the official Qwen-VL training script. The training was performed on two NVIDIA A100 GPUs using a distributed setup facilitated by DeepSpeed with the Zero-3 optimization strategy. Key hyperparameters for the fine-tuning process were as follows: we trained for 10 epochs with a learning rate of 2e-6 and a cosine learning rate scheduler with a warmup ratio of 0.03. The effective batch size was 8, achieved with a per-device batch size of 2 and 2 gradient accumulation steps. We utilized BF16 for mixed-precision training and enabled gradient checkpointing to manage memory usage. The model's maximum context length was set to 8192 tokens.

\subsubsection{Main Experiment Results}
This section provides a detailed, per-model breakdown of results on the Res-Bench dataset. We evaluated a diverse suite of eight leading MLLMs, including proprietary models like Gemini 1.5 Pro \cite{gemini} and GPT-4o \cite{gpt4o}, as well as open-source models such as Qwen2.5-VL \cite{qwen2vl}, LLaVA-OneVision \cite{llava-ov}, InternVL2.5 \cite{InternVL2.5}, DeepSeek-VL\cite{deepseekvl}, MiniCPM-o-2.6 \cite{minicpm} and mPLUG-Owl3 \cite{mplug}.
For each of the evaluated models, we present the full results across all 15 sub-tasks, including accuracy at each of the 12 resolution levels and the four overall robustness metrics. These tables offer the most granular view of model performance, complementing the aggregated analyses in the main text. The abbreviations for sub-tasks are defined as follows: IS (Image Style), ISU (Image Scene Understanding),  IA (Instance Attribute), IL (Instance Location), IC (Instance Counting), CIC (Cross-instance Comparison), SRR (Spatial Relation Reasoning), CDR (Chart \& Diagram Reasoning), ST (Science \& Technology), FC (Function Calculation), Geo (Geometry), SA (Statistical Analysis), DVQA (Doc-oriented VQA), KIE (Key Information Extraction), STU (Scene Text Understanding).

\begin{table*}[htbp]
\centering

\label{tab:full_l2_results_gemini}
\small 
\setlength{\tabcolsep}{1mm} 
\resizebox{\textwidth}{!}{\begin{tabular}{lcccccccccccccccc}
\toprule
\multirow{2}{*}{\textbf{Sub-Tasks}} & \multicolumn{12}{c}{\textbf{Accuracy at Each Resolution}} & \multirow{2}{*}{\textbf{$Acc_{\text{avg}}$ $\uparrow$}} & \multirow{2}{*}{\textbf{$\rho$ $\uparrow$}} & \multirow{2}{*}{\textbf{$ACE \downarrow$}} & \multirow{2}{*}{\textbf{$RCE \downarrow$}}\\
 & 112 & 224 & 336 & 448 & 560 & 672 & 784 & 896 & 1008 & 1120 & 1232 & 1344 &  \\
\midrule
IS & 0.670 & 0.720 & 0.720 & 0.680 & 0.720 & 0.700 & 0.730 & 0.700 & 0.720 & 0.740 & 0.730 & 0.710 & 0.712 & 0.447 & 0.280 & 0.393 \\
ISU & 0.830 & 0.830 & 0.800 & 0.790 & 0.790 & 0.790 & 0.790 & 0.810 & 0.830 & 0.810 & 0.830 & 0.810 & 0.809 & 0.106 & 0.140 & 0.173 \\
\midrule
IA & 0.244 & 0.220 & 0.341 & 0.463 & 0.488 & 0.488 & 0.537 & 0.585 & 0.561 & 0.488 & 0.561 & 0.488 & 0.455 & 0.752 & 0.634 & 1.393 \\
IL & 0.500 & 0.571 & 0.524 & 0.524 & 0.500 & 0.476 & 0.524 & 0.500 & 0.500 & 0.476 & 0.548 & 0.548 & 0.516 & -0.014 & 0.333 & 0.646 \\
IC & 0.410 & 0.487 & 0.462 & 0.470 & 0.462 & 0.470 & 0.453 & 0.487 & 0.436 & 0.504 & 0.513 & 0.504 & 0.472 & 0.577 & 0.316 & 0.671 \\
\midrule
CIC & 0.628 & 0.718 & 0.744 & 0.744 & 0.782 & 0.808 & 0.744 & 0.821 & 0.795 & 0.731 & 0.769 & 0.795 & 0.756 & 0.540 & 0.474 & 0.627 \\
SRR & 0.434 & 0.500 & 0.500 & 0.639 & 0.574 & 0.574 & 0.598 & 0.598 & 0.607 & 0.590 & 0.623 & 0.598 & 0.570 & 0.643 & 0.377 & 0.662 \\
\midrule
CDR & 0.496 & 0.487 & 0.479 & 0.454 & 0.487 & 0.479 & 0.471 & 0.479 & 0.445 & 0.454 & 0.496 & 0.471 & 0.475 & -0.383 & 0.210 & 0.442 \\
ST & 0.222 & 0.210 & 0.222 & 0.469 & 0.457 & 0.494 & 0.469 & 0.444 & 0.494 & 0.469 & 0.444 & 0.444 & 0.403 & 0.402 & 0.469 & 1.163 \\
\midrule
FC & 0.369 & 0.393 & 0.405 & 0.417 & 0.440 & 0.452 & 0.417 & 0.452 & 0.440 & 0.464 & 0.429 & 0.452 & 0.428 & 0.760 & 0.250 & 0.585 \\
Geo & 0.304 & 0.337 & 0.370 & 0.391 & 0.391 & 0.391 & 0.380 & 0.380 & 0.370 & 0.359 & 0.348 & 0.391 & 0.368 & 0.236 & 0.174 & 0.473 \\
SA & 0.292 & 0.250 & 0.417 & 0.542 & 0.417 & 0.458 & 0.458 & 0.500 & 0.500 & 0.458 & 0.417 & 0.417 & 0.427 & 0.338 & 0.625 & 1.463 \\
\midrule
DVQA & 0.156 & 0.284 & 0.556 & 0.689 & 0.847 & 0.840 & 0.842 & 0.853 & 0.837 & 0.862 & 0.844 & 0.851 & 0.705 & 0.797 & 0.776 & 1.101 \\
KIE & 0.031 & 0.171 & 0.390 & 0.613 & 0.697 & 0.733 & 0.786 & 0.688 & 0.768 & 0.715 & 0.736 & 0.707 & 0.586 & 0.727 & 1.035 & 1.765 \\
STU & 0.236 & 0.546 & 0.620 & 0.649 & 0.677 & 0.689 & 0.726 & 0.700 & 0.714 & 0.717 & 0.727 & 0.705 & 0.642 & 0.881 & 0.564 & 0.879 \\
\bottomrule
\end{tabular}
}
\caption{Detailed performance and robustness results for Gemini1.5-Pro on all 15 sub-tasks.}
\end{table*}

\begin{table*}[htbp]
\centering

\label{tab:full_l2_results_gpt4o}
\small 
\setlength{\tabcolsep}{1mm} 
\resizebox{\textwidth}{!}{\begin{tabular}{lcccccccccccccccc}
\toprule
\multirow{2}{*}{\textbf{Sub-Task}} & \multicolumn{12}{c}{\textbf{Accuracy at Each Resolution}} & \multirow{2}{*}{\textbf{$Acc_{avg} \uparrow$}} &\multirow{2}{*}{\textbf{$\rho \uparrow$}} & \multirow{2}{*}{\textbf{$ACE \downarrow$}} & \multirow{2}{*}{\textbf{$RCE \downarrow$}} \\
 & 112 & 224 & 336 & 448 & 560 & 672 & 784 & 896 & 1008 & 1120 & 1232 & 1344 & \\
\midrule
IS & 0.710 & 0.730 & 0.800 & 0.840 & 0.780 & 0.800 & 0.770 & 0.780 & 0.800 & 0.740 & 0.770 & 0.790 & 0.776 & 0.124 & 0.380 & 0.490 \\
ISU & 0.830 & 0.810 & 0.800 & 0.770 & 0.790 & 0.790 & 0.800 & 0.800 & 0.770 & 0.810 & 0.800 & 0.800 & 0.798 & -0.175 & 0.170 & 0.213 \\
\midrule
IA & 0.366 & 0.390 & 0.415 & 0.463 & 0.512 & 0.561 & 0.512 & 0.561 & 0.634 & 0.512 & 0.610 & 0.585 & 0.510 & 0.868 & 0.610 & 1.195 \\
IL & 0.500 & 0.476 & 0.524 & 0.381 & 0.476 & 0.429 & 0.524 & 0.524 & 0.429 & 0.476 & 0.452 & 0.452 & 0.470 & -0.256 & 0.619 & 1.316 \\
IC & 0.402 & 0.470 & 0.410 & 0.419 & 0.462 & 0.419 & 0.462 & 0.444 & 0.521 & 0.504 & 0.521 & 0.479 & 0.459 & 0.752 & 0.436 & 0.949 \\
\midrule
CIC & 0.718 & 0.808 & 0.821 & 0.821 & 0.821 & 0.846 & 0.872 & 0.897 & 0.859 & 0.859 & 0.859 & 0.910 & 0.841 & 0.880 & 0.269 & 0.320 \\
SRR & 0.361 & 0.426 & 0.410 & 0.418 & 0.402 & 0.434 & 0.402 & 0.344 & 0.443 & 0.418 & 0.426 & 0.418 & 0.408 & 0.279 & 0.369 & 0.903 \\
\midrule
CDR & 0.395 & 0.412 & 0.420 & 0.387 & 0.370 & 0.395 & 0.412 & 0.412 & 0.429 & 0.403 & 0.403 & 0.403 & 0.403 & 0.160 & 0.160 & 0.396 \\
ST & 0.407 & 0.432 & 0.420 & 0.407 & 0.444 & 0.420 & 0.457 & 0.407 & 0.420 & 0.444 & 0.383 & 0.494 & 0.428 & 0.217 & 0.407 & 0.952 \\
\midrule
FC & 0.310 & 0.369 & 0.345 & 0.369 & 0.381 & 0.417 & 0.393 & 0.393 & 0.369 & 0.405 & 0.369 & 0.393 & 0.376 & 0.552 & 0.298 & 0.792 \\
Geo & 0.293 & 0.228 & 0.359 & 0.293 & 0.250 & 0.293 & 0.283 & 0.283 & 0.293 & 0.261 & 0.315 & 0.315 & 0.289 & 0.214 & 0.457 & 1.580 \\
SA & 0.292 & 0.208 & 0.208 & 0.333 & 0.292 & 0.292 & 0.292 & 0.375 & 0.458 & 0.458 & 0.458 & 0.375 & 0.330 & 0.880 & 0.417 & 1.263 \\
\midrule
DVQA & 0.209 & 0.309 & 0.375 & 0.597 & 0.674 & 0.713 & 0.802 & 0.818 & 0.856 & 0.865 & 0.839 & 0.844 & 0.658 & 0.944 & 0.686 & 1.042 \\
KIE & 0.016 & 0.187 & 0.320 & 0.457 & 0.533 & 0.557 & 0.633 & 0.691 & 0.663 & 0.684 & 0.675 & 0.710 & 0.511 & 0.951 & 0.769 & 1.507 \\
STU & 0.249 & 0.483 & 0.551 & 0.560 & 0.593 & 0.594 & 0.623 & 0.639 & 0.616 & 0.635 & 0.638 & 0.616 & 0.566 & 0.839 & 0.456 & 0.805 \\
\bottomrule
\end{tabular}
} %
\caption{Detailed performance and robustness results for GPT-4o on all 15 sub-tasks.}
\end{table*}

%
%
\begin{table*}[htbp]
\centering

\label{tab:full_l2_results}
\small 
\setlength{\tabcolsep}{1mm}
\resizebox{\textwidth}{!}{\begin{tabular}{lcccccccccccccccc}
\toprule
\multirow{2}{*}{\textbf{Sub-Tasks}} & \multicolumn{12}{c}{\textbf{Accuracy at Each Resolution}} & \multirow{2}{*}{\textbf{$Acc_{\text{avg}}\uparrow$ }} & \multirow{2}{*}{\textbf{$\rho \uparrow$}} & \multirow{2}{*}{\textbf{$ACE \downarrow$}} & \multirow{2}{*}{\textbf{$RCE \downarrow$}}\\
 & 112 & 224 & 336 & 448 & 560 & 672 & 784 & 896 & 1008 & 1120 & 1232 & 1344 &  \\
\midrule
IS & 0.660 & 0.690 & 0.680 & 0.710 & 0.690 & 0.690 & 0.680 & 0.670 & 0.690 & 0.650 & 0.690 & 0.660 & 0.680 & -0.237 & 0.240 & 0.353 \\
ISU & 0.740 & 0.800 & 0.830 & 0.810 & 0.840 & 0.810 & 0.850 & 0.820 & 0.840 & 0.820 & 0.810 & 0.810 & 0.815 & 0.282 & 0.290 & 0.356 \\
\midrule
IA & 0.415 & 0.268 & 0.390 & 0.439 & 0.488 & 0.537 & 0.610 & 0.561 & 0.610 & 0.585 & 0.683 & 0.707 & 0.524 & 0.939 & 0.732 & 1.395 \\
IL & 0.357 & 0.476 & 0.500 & 0.524 & 0.524 & 0.571 & 0.571 & 0.619 & 0.548 & 0.595 & 0.548 & 0.452 & 0.524 & 0.457 & 0.524 & 1.000 \\
IC & 0.316 & 0.402 & 0.513 & 0.410 & 0.504 & 0.513 & 0.521 & 0.496 & 0.598 & 0.547 & 0.564 & 0.581 & 0.497 & 0.851 & 0.624 & 1.255 \\
\midrule
CIC & 0.436 & 0.641 & 0.692 & 0.679 & 0.705 & 0.782 & 0.718 & 0.782 & 0.782 & 0.769 & 0.782 & 0.744 & 0.709 & 0.783 & 0.564 & 0.795 \\
SRR & 0.418 & 0.533 & 0.590 & 0.598 & 0.631 & 0.664 & 0.648 & 0.664 & 0.705 & 0.664 & 0.713 & 0.680 & 0.626 & 0.937 & 0.443 & 0.707 \\
\midrule
CDR & 0.437 & 0.471 & 0.496 & 0.521 & 0.563 & 0.513 & 0.529 & 0.597 & 0.563 & 0.513 & 0.605 & 0.571 & 0.532 & 0.779 & 0.471 & 0.885 \\
ST & 0.358 & 0.321 & 0.309 & 0.370 & 0.333 & 0.346 & 0.333 & 0.383 & 0.358 & 0.407 & 0.383 & 0.383 & 0.357 & 0.700 & 0.321 & 0.899 \\
\midrule
FC & 0.310 & 0.357 & 0.357 & 0.381 & 0.393 & 0.393 & 0.417 & 0.405 & 0.369 & 0.345 & 0.345 & 0.369 & 0.370 & 0.099 & 0.202 & 0.547 \\
Geo & 0.272 & 0.359 & 0.380 & 0.413 & 0.424 & 0.402 & 0.370 & 0.424 & 0.424 & 0.380 & 0.424 & 0.424 & 0.391 & 0.665 & 0.348 & 0.889 \\
SA & 0.125 & 0.167 & 0.250 & 0.333 & 0.292 & 0.292 & 0.333 & 0.333 & 0.292 & 0.375 & 0.375 & 0.333 & 0.292 & 0.805 & 0.458 & 1.571 \\
\midrule
DVQA & 0.094 & 0.180 & 0.361 & 0.565 & 0.695 & 0.757 & 0.803 & 0.802 & 0.857 & 0.808 & 0.826 & 0.832 & 0.632 & 0.951 & 0.828 & 1.310 \\
KIE & 0.003 & 0.200 & 0.386 & 0.540 & 0.663 & 0.695 & 0.736 & 0.774 & 0.750 & 0.744 & 0.746 & 0.762 & 0.582 & 0.909 & 0.829 & 1.399 \\
STU & 0.146 & 0.394 & 0.518 & 0.591 & 0.618 & 0.631 & 0.667 & 0.678 & 0.664 & 0.669 & 0.670 & 0.674 & 0.577 & 0.916 & 0.555 & 0.963 \\
\bottomrule
\end{tabular}}
\caption{Detailed performance and robustness results for Qwen2.5-VL on all 15 sub-tasks.}
\end{table*}

\begin{table*}[htbp]
\centering
\label{tab:full_l2_results_kimi}
\small 
\setlength{\tabcolsep}{1mm} 
\resizebox{\textwidth}{!}{\begin{tabular}{lcccccccccccccccc}
\toprule
\multirow{2}{*}{\textbf{Sub-Task}} & \multicolumn{12}{c}{\textbf{Accuracy at Each Resolution}} &\multirow{2}{*}{$Acc_{avg} \uparrow$} & \multirow{2}{*}{\textbf{$\rho \uparrow$}} & \multirow{2}{*}{\textbf{$ACE \downarrow$}} & \multirow{2}{*}{\textbf{$RCE \downarrow$}}\\
 & 112 & 224 & 336 & 448 & 560 & 672 & 784 & 896 & 1008 & 1120 & 1232 & 1344 &  \\
\midrule
IS & 0.690 & 0.660 & 0.650 & 0.630 & 0.660 & 0.710 & 0.680 & 0.680 & 0.660 & 0.670 & 0.700 & 0.670 & 0.672 & 0.286 & 0.260 & 0.387 \\
ISU & 0.750 & 0.790 & 0.800 & 0.840 & 0.830 & 0.840 & 0.820 & 0.850 & 0.830 & 0.830 & 0.800 & 0.820 & 0.817 & 0.315 & 0.230 & 0.282 \\
\midrule
IA & 0.244 & 0.268 & 0.415 & 0.463 & 0.512 & 0.561 & 0.610 & 0.634 & 0.610 & 0.610 & 0.659 & 0.634 & 0.518 & 0.945 & 0.488 & 0.941 \\
IL & 0.524 & 0.571 & 0.595 & 0.548 & 0.548 & 0.548 & 0.571 & 0.524 & 0.571 & 0.524 & 0.524 & 0.548 & 0.550 & -0.296 & 0.310 & 0.563 \\
IC & 0.410 & 0.436 & 0.547 & 0.513 & 0.538 & 0.538 & 0.521 & 0.530 & 0.564 & 0.547 & 0.538 & 0.513 & 0.516 & 0.406 & 0.308 & 0.596 \\
\midrule
CIC & 0.615 & 0.667 & 0.769 & 0.821 & 0.821 & 0.821 & 0.846 & 0.795 & 0.769 & 0.808 & 0.821 & 0.795 & 0.779 & 0.364 & 0.385 & 0.494 \\
SRR & 0.426 & 0.484 & 0.484 & 0.557 & 0.582 & 0.574 & 0.541 & 0.557 & 0.566 & 0.541 & 0.549 & 0.582 & 0.537 & 0.532 & 0.287 & 0.534 \\
\midrule
CDR & 0.370 & 0.303 & 0.328 & 0.378 & 0.403 & 0.395 & 0.395 & 0.487 & 0.403 & 0.395 & 0.403 & 0.462 & 0.394 & 0.787 & 0.429 & 1.089 \\
ST & 0.346 & 0.370 & 0.370 & 0.358 & 0.395 & 0.383 & 0.395 & 0.395 & 0.383 & 0.407 & 0.432 & 0.383 & 0.385 & 0.735 & 0.210 & 0.545 \\
\midrule
FC & 0.167 & 0.250 & 0.202 & 0.262 & 0.214 & 0.226 & 0.262 & 0.298 & 0.333 & 0.357 & 0.310 & 0.381 & 0.272 & 0.893 & 0.500 & 1.839 \\
Geo & 0.152 & 0.141 & 0.152 & 0.163 & 0.196 & 0.098 & 0.109 & 0.152 & 0.130 & 0.196 & 0.163 & 0.196 & 0.154 & 0.316 & 0.370 & 2.400 \\
SA & 0.208 & 0.292 & 0.333 & 0.375 & 0.500 & 0.417 & 0.417 & 0.500 & 0.500 & 0.500 & 0.458 & 0.458 & 0.413 & 0.736 & 0.500 & 1.210 \\
\midrule
DVQA & 0.059 & 0.145 & 0.343 & 0.606 & 0.749 & 0.807 & 0.841 & 0.886 & 0.886 & 0.880 & 0.886 & 0.905 & 0.666 & 0.965 & 0.858 & 1.287 \\
KIE & 0.012 & 0.201 & 0.428 & 0.554 & 0.658 & 0.686 & 0.721 & 0.738 & 0.738 & 0.757 & 0.759 & 0.766 & 0.585 & 0.993 & 0.753 & 1.288 \\
STU & 0.119 & 0.350 & 0.473 & 0.557 & 0.580 & 0.598 & 0.636 & 0.643 & 0.636 & 0.666 & 0.645 & 0.624 & 0.544 & 0.881 & 0.603 & 1.109 \\
\bottomrule
\end{tabular}
} %
\caption{Detailed performance and robustness results for Kimi-VL on all 15 sub-tasks.}
\end{table*}

\begin{table*}[htbp]
\centering

\label{tab:full_l2_results_llava}
\small 
\setlength{\tabcolsep}{1mm}
\resizebox{\textwidth}{!}{\begin{tabular}{lcccccccccccccccc}
\toprule
\multirow{2}{*}{\textbf{Sub-Task}} & \multicolumn{12}{c}{\textbf{Accuracy at Each Resolution}} & \multirow{2}{*}{$Acc_{avg} \uparrow$} & \multirow{2}{*}{\textbf{$\rho \uparrow$}} & \multirow{2}{*}{\textbf{$ACE \downarrow$}} & \multirow{2}{*}{\textbf{$RCE \downarrow$}}  \\
 & 112 & 224 & 336 & 448 & 560 & 672 & 784 & 896 & 1008 & 1120 & 1232 & 1344 &  \\
\midrule
IS & 0.630 & 0.700 & 0.750 & 0.680 & 0.640 & 0.640 & 0.640 & 0.610 & 0.590 & 0.640 & 0.560 & 0.580 & 0.638 & -0.712 & 0.430 & 0.674 \\
ISU & 0.810 & 0.810 & 0.820 & 0.850 & 0.840 & 0.840 & 0.820 & 0.830 & 0.820 & 0.830 & 0.850 & 0.830 & 0.829 & 0.431 & 0.140 & 0.169 \\
\midrule
IA & 0.317 & 0.439 & 0.439 & 0.463 & 0.585 & 0.610 & 0.683 & 0.683 & 0.707 & 0.780 & 0.707 & 0.659 & 0.589 & 0.879 & 0.585 & 0.993 \\
IL & 0.500 & 0.524 & 0.476 & 0.500 & 0.524 & 0.500 & 0.571 & 0.548 & 0.500 & 0.548 & 0.571 & 0.595 & 0.530 & 0.698 & 0.381 & 0.719 \\
IC & 0.479 & 0.598 & 0.590 & 0.564 & 0.564 & 0.530 & 0.556 & 0.538 & 0.547 & 0.530 & 0.530 & 0.530 & 0.546 & -0.428 & 0.256 & 0.469 \\
\midrule
CIC & 0.538 & 0.718 & 0.718 & 0.692 & 0.718 & 0.731 & 0.654 & 0.667 & 0.641 & 0.718 & 0.679 & 0.667 & 0.678 & -0.150 & 0.487 & 0.718 \\
SRR & 0.467 & 0.549 & 0.582 & 0.590 & 0.631 & 0.590 & 0.623 & 0.623 & 0.656 & 0.648 & 0.656 & 0.639 & 0.605 & 0.896 & 0.303 & 0.502 \\
\midrule
CDR & 0.521 & 0.513 & 0.513 & 0.513 & 0.513 & 0.496 & 0.496 & 0.504 & 0.521 & 0.521 & 0.504 & 0.538 & 0.513 & 0.101 & 0.101 & 0.197 \\
ST & 0.321 & 0.346 & 0.358 & 0.333 & 0.346 & 0.333 & 0.333 & 0.358 & 0.333 & 0.358 & 0.358 & 0.346 & 0.344 & 0.395 & 0.173 & 0.503 \\
\midrule
FC & 0.262 & 0.250 & 0.238 & 0.250 & 0.214 & 0.274 & 0.238 & 0.238 & 0.262 & 0.250 & 0.286 & 0.226 & 0.249 & -0.007 & 0.298 & 1.195 \\
Geo & 0.326 & 0.326 & 0.304 & 0.326 & 0.315 & 0.304 & 0.304 & 0.326 & 0.337 & 0.359 & 0.315 & 0.337 & 0.323 & 0.360 & 0.185 & 0.571 \\
SA & 0.250 & 0.333 & 0.333 & 0.333 & 0.333 & 0.292 & 0.292 & 0.375 & 0.375 & 0.375 & 0.375 & 0.375 & 0.337 & 0.767 & 0.208 & 0.619 \\
\midrule
DVQA & 0.151 & 0.265 & 0.392 & 0.525 & 0.683 & 0.724 & 0.764 & 0.758 & 0.782 & 0.787 & 0.746 & 0.751 & 0.611 & 0.825 & 0.694 & 1.136 \\
KIE & 0.008 & 0.088 & 0.142 & 0.221 & 0.270 & 0.293 & 0.339 & 0.351 & 0.379 & 0.368 & 0.360 & 0.357 & 0.265 & 0.930 & 0.393 & 1.484 \\
STU & 0.200 & 0.419 & 0.473 & 0.523 & 0.529 & 0.526 & 0.538 & 0.565 & 0.564 & 0.565 & 0.565 & 0.561 & 0.502 & 0.909 & 0.376 & 0.749 \\
\bottomrule
\end{tabular}
} %
\caption{Detailed performance and robustness results for LLaVA-OneVision on all 15 sub-tasks.}
\end{table*}

\begin{table*}[htbp]
\centering
\label{tab:full_l2_results_mplug}
\small 
\setlength{\tabcolsep}{1mm}
\resizebox{\textwidth}{!}{\begin{tabular}{lcccccccccccccccc}
\toprule
\multirow{2}{*}{\textbf{Sub-Task}} & \multicolumn{12}{c|}{\textbf{Accuracy at Each Resolution}} & \multirow{2}{*}{$Acc_{avg} \uparrow$} & \multirow{2}{*}{\textbf{$\rho \uparrow$}} & \multirow{2}{*}{\textbf{$ACE \downarrow$}} & \multirow{2}{*}{\textbf{$RCE \downarrow$}}\\
 & 112 & 224 & 336 & 448 & 560 & 672 & 784 & 896 & 1008 & 1120 & 1232 & 1344 &  \\
\midrule
IS & 0.390 & 0.090 & 0.160 & 0.110 & 0.100 & 0.270 & 0.250 & 0.310 & 0.340 & 0.340 & 0.330 & 0.360 & 0.254 & 0.480 & 0.750 & 2.951 \\
ISU & 0.870 & 0.880 & 0.840 & 0.860 & 0.870 & 0.850 & 0.850 & 0.830 & 0.850 & 0.840 & 0.840 & 0.840 & 0.852 & -0.690 & 0.150 & 0.176 \\
\midrule
IA & 0.220 & 0.317 & 0.366 & 0.415 & 0.341 & 0.463 & 0.488 & 0.463 & 0.439 & 0.488 & 0.415 & 0.415 & 0.402 & 0.618 & 0.585 & 1.455 \\
IL & 0.476 & 0.524 & 0.524 & 0.571 & 0.476 & 0.524 & 0.548 & 0.476 & 0.500 & 0.524 & 0.524 & 0.548 & 0.518 & 0.209 & 0.405 & 0.782 \\
IC & 0.410 & 0.444 & 0.427 & 0.427 & 0.453 & 0.462 & 0.470 & 0.470 & 0.487 & 0.462 & 0.470 & 0.453 & 0.453 & 0.715 & 0.162 & 0.358 \\
\midrule
CIC & 0.577 & 0.590 & 0.641 & 0.603 & 0.590 & 0.590 & 0.590 & 0.641 & 0.654 & 0.628 & 0.654 & 0.641 & 0.616 & 0.682 & 0.244 & 0.395 \\
SRR & 0.189 & 0.320 & 0.426 & 0.393 & 0.320 & 0.352 & 0.369 & 0.377 & 0.393 & 0.402 & 0.402 & 0.369 & 0.359 & 0.444 & 0.459 & 1.278 \\
\midrule
CDR & 0.395 & 0.412 & 0.387 & 0.378 & 0.361 & 0.378 & 0.378 & 0.378 & 0.370 & 0.370 & 0.378 & 0.378 & 0.380 & -0.580 & 0.101 & 0.265 \\
ST & 0.222 & 0.333 & 0.321 & 0.370 & 0.346 & 0.333 & 0.358 & 0.346 & 0.346 & 0.346 & 0.346 & 0.333 & 0.333 & 0.358 & 0.259 & 0.778 \\
\midrule
FC & 0.226 & 0.226 & 0.214 & 0.202 & 0.226 & 0.202 & 0.202 & 0.214 & 0.238 & 0.214 & 0.226 & 0.226 & 0.218 & 0.140 & 0.143 & 0.655 \\
Geo & 0.185 & 0.120 & 0.120 & 0.120 & 0.109 & 0.109 & 0.109 & 0.120 & 0.130 & 0.120 & 0.130 & 0.130 & 0.125 & 0.169 & 0.120 & 0.957 \\
SA & 0.292 & 0.333 & 0.333 & 0.250 & 0.292 & 0.292 & 0.292 & 0.292 & 0.250 & 0.250 & 0.250 & 0.292 & 0.285 & -0.550 & 0.250 & 0.878 \\
\midrule
DVQA & 0.033 & 0.141 & 0.232 & 0.266 & 0.316 & 0.320 & 0.328 & 0.310 & 0.344 & 0.319 & 0.334 & 0.319 & 0.272 & 0.778 & 0.404 & 1.487 \\
KIE & 0.009 & 0.058 & 0.144 & 0.188 & 0.215 & 0.204 & 0.196 & 0.207 & 0.201 & 0.205 & 0.192 & 0.196 & 0.168 & 0.510 & 0.262 & 1.561 \\
STU & 0.215 & 0.391 & 0.428 & 0.462 & 0.437 & 0.453 & 0.457 & 0.456 & 0.441 & 0.450 & 0.447 & 0.446 & 0.424 & 0.420 & 0.321 & 0.757 \\
\bottomrule
\end{tabular}
} %
\caption{Detailed performance and robustness results for mPLUG-Owl3 on all 15 sub-tasks.}
\end{table*}

\begin{table*}[htbp]
\centering
\label{tab:full_l2_results_internvl}
\small
\setlength{\tabcolsep}{1mm} 
\resizebox{\textwidth}{!}{\begin{tabular}{lcccccccccccccccc}
\toprule
\multirow{2}{*}{\textbf{Sub-Task}} & \multicolumn{12}{c}{\textbf{Accuracy at Each Resolution}} & \multirow{2}{*}{$Acc_{avg} \uparrow$} & \multirow{2}{*}{\textbf{$\rho \uparrow$}} & \multirow{2}{*}{\textbf{$ACE \downarrow$}} & \multirow{2}{*}{\textbf{$RCE \downarrow$}} \\
 & 112 & 224 & 336 & 448 & 560 & 672 & 784 & 896 & 1008 & 1120 & 1232 & 1344 & \\
\midrule
IS & 0.640 & 0.740 & 0.780 & 0.770 & 0.740 & 0.740 & 0.760 & 0.740 & 0.720 & 0.710 & 0.700 & 0.720 & 0.730 & -0.371 & 0.280 & 0.384 \\
ISU & 0.760 & 0.800 & 0.790 & 0.800 & 0.770 & 0.820 & 0.790 & 0.790 & 0.800 & 0.820 & 0.810 & 0.820 & 0.798 & 0.629 & 0.220 & 0.276 \\
\midrule
IA & 0.415 & 0.537 & 0.561 & 0.610 & 0.561 & 0.561 & 0.585 & 0.634 & 0.610 & 0.610 & 0.634 & 0.585 & 0.575 & 0.733 & 0.415 & 0.721 \\
IL & 0.476 & 0.476 & 0.452 & 0.476 & 0.500 & 0.500 & 0.476 & 0.476 & 0.476 & 0.452 & 0.476 & 0.476 & 0.476 & -0.084 & 0.143 & 0.300 \\
IC & 0.376 & 0.496 & 0.513 & 0.487 & 0.521 & 0.521 & 0.521 & 0.538 & 0.504 & 0.521 & 0.513 & 0.521 & 0.503 & 0.559 & 0.282 & 0.561 \\
\midrule
CIC & 0.667 & 0.744 & 0.821 & 0.795 & 0.808 & 0.808 & 0.808 & 0.833 & 0.795 & 0.872 & 0.872 & 0.859 & 0.807 & 0.770 & 0.346 & 0.429 \\
SRR & 0.484 & 0.525 & 0.549 & 0.607 & 0.615 & 0.623 & 0.607 & 0.598 & 0.615 & 0.623 & 0.631 & 0.598 & 0.589 & 0.616 & 0.230 & 0.389 \\
\midrule
CDR & 0.370 & 0.387 & 0.403 & 0.420 & 0.420 & 0.412 & 0.403 & 0.395 & 0.395 & 0.395 & 0.395 & 0.387 & 0.398 & 0.097 & 0.084 & 0.211 \\
ST & 0.309 & 0.309 & 0.309 & 0.333 & 0.309 & 0.333 & 0.284 & 0.333 & 0.333 & 0.309 & 0.321 & 0.333 & 0.318 & 0.385 & 0.222 & 0.699 \\
\midrule
FC & 0.310 & 0.310 & 0.310 & 0.333 & 0.345 & 0.310 & 0.321 & 0.333 & 0.298 & 0.310 & 0.310 & 0.310 & 0.316 & -0.188 & 0.143 & 0.451 \\
Geo & 0.326 & 0.272 & 0.261 & 0.283 & 0.261 & 0.261 & 0.293 & 0.250 & 0.283 & 0.272 & 0.272 & 0.272 & 0.275 & -0.122 & 0.228 & 0.829 \\
SA & 0.250 & 0.250 & 0.375 & 0.375 & 0.375 & 0.333 & 0.375 & 0.333 & 0.333 & 0.375 & 0.375 & 0.375 & 0.344 & 0.453 & 0.292 & 0.848 \\
\midrule
DVQA & 0.151 & 0.195 & 0.313 & 0.602 & 0.740 & 0.746 & 0.783 & 0.779 & 0.783 & 0.786 & 0.789 & 0.808 & 0.621 & 0.993 & 0.693 & 1.117 \\
KIE & 0.008 & 0.127 & 0.316 & 0.477 & 0.637 & 0.663 & 0.683 & 0.693 & 0.721 & 0.715 & 0.715 & 0.689 & 0.537 & 0.902 & 0.749 & 1.396 \\
STU & 0.170 & 0.352 & 0.475 & 0.511 & 0.526 & 0.556 & 0.567 & 0.595 & 0.572 & 0.585 & 0.580 & 0.577 & 0.505 & 0.902 & 0.467 & 0.924 \\
\bottomrule
\end{tabular}
} %
\caption{Detailed performance and robustness results for InternVL-2.5 on all 15 sub-tasks.}
\end{table*}

\begin{table*}[htbp]
\centering

\label{tab:full_l2_results_minicpm}
\small 
\setlength{\tabcolsep}{1mm}
\resizebox{\textwidth}{!}{\begin{tabular}{lcccccccccccccccc}
\toprule
\multirow{2}{*}{\textbf{Sub-Task}} & \multicolumn{12}{c}{\textbf{Accuracy at Each Resolution}} &  \multirow{2}{*}{$Acc_{avg} \uparrow$} &  \multirow{2}{*}{\textbf{$\rho \uparrow$}} &  \multirow{2}{*}{\textbf{$ACE \downarrow$}} &  \multirow{2}{*}{\textbf{$RCE \downarrow$}} \\
 & 112 & 224 & 336 & 448 & 560 & 672 & 784 & 896 & 1008 & 1120 & 1232 & 1344 & \\
\midrule
IS & 0.700 & 0.770 & 0.700 & 0.750 & 0.770 & 0.780 & 0.790 & 0.810 & 0.770 & 0.780 & 0.810 & 0.780 & 0.768 & 0.737 & 0.360 & 0.469 \\
ISU & 0.790 & 0.810 & 0.810 & 0.800 & 0.810 & 0.810 & 0.810 & 0.820 & 0.820 & 0.830 & 0.860 & 0.860 & 0.819 & 0.917 & 0.090 & 0.110 \\
\midrule
IA & 0.268 & 0.366 & 0.390 & 0.390 & 0.537 & 0.488 & 0.463 & 0.537 & 0.683 & 0.683 & 0.683 & 0.707 & 0.516 & 0.954 & 0.585 & 1.134 \\
IL & 0.381 & 0.619 & 0.571 & 0.571 & 0.500 & 0.571 & 0.595 & 0.595 & 0.619 & 0.571 & 0.548 & 0.548 & 0.558 & 0.021 & 0.548 & 0.982 \\
IC & 0.453 & 0.496 & 0.530 & 0.538 & 0.547 & 0.496 & 0.530 & 0.556 & 0.573 & 0.581 & 0.573 & 0.607 & 0.540 & 0.886 & 0.274 & 0.507 \\
\midrule
CIC & 0.551 & 0.641 & 0.731 & 0.705 & 0.756 & 0.782 & 0.744 & 0.718 & 0.756 & 0.718 & 0.692 & 0.744 & 0.712 & 0.334 & 0.500 & 0.703 \\
SRR & 0.467 & 0.549 & 0.607 & 0.623 & 0.648 & 0.639 & 0.664 & 0.639 & 0.648 & 0.697 & 0.746 & 0.738 & 0.639 & 0.926 & 0.352 & 0.552 \\
\midrule
CDR & 0.202 & 0.160 & 0.218 & 0.277 & 0.286 & 0.303 & 0.286 & 0.303 & 0.277 & 0.277 & 0.277 & 0.311 & 0.265 & 0.614 & 0.277 & 1.048 \\
ST & 0.235 & 0.222 & 0.222 & 0.259 & 0.247 & 0.210 & 0.235 & 0.272 & 0.296 & 0.272 & 0.259 & 0.247 & 0.248 & 0.582 & 0.235 & 0.946 \\
\midrule
FC & 0.226 & 0.238 & 0.238 & 0.262 & 0.298 & 0.286 & 0.274 & 0.274 & 0.298 & 0.321 & 0.298 & 0.274 & 0.274 & 0.728 & 0.190 & 0.696 \\
Geo & 0.217 & 0.337 & 0.283 & 0.315 & 0.337 & 0.359 & 0.315 & 0.293 & 0.293 & 0.337 & 0.326 & 0.370 & 0.315 & 0.424 & 0.413 & 1.310 \\
SA & 0.167 & 0.167 & 0.250 & 0.500 & 0.333 & 0.333 & 0.417 & 0.375 & 0.417 & 0.500 & 0.417 & 0.375 & 0.354 & 0.638 & 0.875 & 2.471 \\
\midrule
DVQA & 0.109 & 0.189 & 0.371 & 0.531 & 0.682 & 0.772 & 0.784 & 0.830 & 0.826 & 0.830 & 0.843 & 0.880 & 0.637 & 0.993 & 0.779 & 1.223 \\
KIE & 0.008 & 0.168 & 0.379 & 0.384 & 0.546 & 0.574 & 0.520 & 0.567 & 0.663 & 0.622 & 0.611 & 0.653 & 0.475 & 0.916 & 0.857 & 1.805 \\
STU & 0.176 & 0.442 & 0.521 & 0.546 & 0.591 & 0.618 & 0.640 & 0.656 & 0.633 & 0.661 & 0.641 & 0.670 & 0.566 & 0.951 & 0.579 & 1.025 \\
\bottomrule
\end{tabular}
} %
\caption{Detailed performance and robustness results for MiniCPM-o-2.6 on all 15 sub-tasks.}
\end{table*}

\subsubsection{Detailed Fine-tuning Experiment Results}

\begin{table*}[htbp]
\centering
\label{tab:full_l2_results_qwen3b_ft}
\small 
\setlength{\tabcolsep}{1mm} 
\resizebox{\textwidth}{!}{\begin{tabular}{lcccccccccccccccc}
\toprule
\multirow{2}{*}{\textbf{Sub-Task}} & \multicolumn{12}{c}{\textbf{Accuracy at Each Resolution}} & \multirow{2}{*}{$Acc_{avg} \uparrow$} & \multirow{2}{*}{\textbf{$\rho \uparrow$}} & \multirow{2}{*}{\textbf{$ACE \downarrow$}} & \multirow{2}{*}{\textbf{$RCE \downarrow$}}\\
 & 112 & 224 & 336 & 448 & 560 & 672 & 784 & 896 & 1008 & 1120 & 1232 & 1344 &  \\
\midrule
IS & 0.480 & 0.420 & 0.490 & 0.540 & 0.470 & 0.550 & 0.590 & 0.560 & 0.570 & 0.550 & 0.570 & 0.610 & 0.533 & 0.835 & 0.490 & 0.919 \\
ISU & 0.690 & 0.750 & 0.780 & 0.820 & 0.810 & 0.810 & 0.850 & 0.860 & 0.830 & 0.820 & 0.810 & 0.840 & 0.806 & 0.677 & 0.270 & 0.335 \\
\midrule
IA & 0.341 & 0.341 & 0.463 & 0.610 & 0.561 & 0.610 & 0.659 & 0.634 & 0.780 & 0.683 & 0.780 & 0.756 & 0.602 & 0.939 & 0.805 & 1.338 \\
IL & 0.262 & 0.381 & 0.548 & 0.452 & 0.500 & 0.524 & 0.571 & 0.571 & 0.595 & 0.571 & 0.571 & 0.548 & 0.508 & 0.767 & 0.571 & 1.125 \\
IC & 0.359 & 0.385 & 0.479 & 0.462 & 0.453 & 0.470 & 0.470 & 0.462 & 0.444 & 0.479 & 0.479 & 0.487 & 0.452 & 0.654 & 0.231 & 0.510 \\
\midrule
CIC & 0.474 & 0.551 & 0.564 & 0.538 & 0.577 & 0.641 & 0.641 & 0.641 & 0.603 & 0.667 & 0.603 & 0.603 & 0.592 & 0.702 & 0.385 & 0.650 \\
SRR & 0.500 & 0.533 & 0.631 & 0.623 & 0.623 & 0.639 & 0.607 & 0.648 & 0.648 & 0.648 & 0.656 & 0.648 & 0.617 & 0.870 & 0.246 & 0.399 \\
\midrule
CDR & 0.395 & 0.429 & 0.454 & 0.471 & 0.504 & 0.504 & 0.521 & 0.513 & 0.513 & 0.521 & 0.496 & 0.504 & 0.485 & 0.678 & 0.176 & 0.364 \\
ST & 0.358 & 0.358 & 0.370 & 0.395 & 0.383 & 0.383 & 0.407 & 0.383 & 0.383 & 0.395 & 0.420 & 0.407 & 0.387 & 0.794 & 0.148 & 0.383 \\
\midrule
FC & 0.262 & 0.250 & 0.262 & 0.298 & 0.298 & 0.298 & 0.298 & 0.286 & 0.298 & 0.286 & 0.274 & 0.262 & 0.278 & 0.309 & 0.190 & 0.686 \\
Geo & 0.261 & 0.261 & 0.337 & 0.315 & 0.315 & 0.304 & 0.326 & 0.315 & 0.326 & 0.326 & 0.380 & 0.337 & 0.318 & 0.633 & 0.250 & 0.786 \\
SA & 0.125 & 0.240 & 0.250 & 0.250 & 0.208 & 0.292 & 0.250 & 0.375 & 0.292 & 0.292 & 0.333 & 0.417 & 0.267 & 0.885 & 0.625 & 2.338 \\
\midrule
DVQA & 0.107 & 0.141 & 0.242 & 0.397 & 0.542 & 0.642 & 0.714 & 0.773 & 0.798 & 0.792 & 0.768 & 0.778 & 0.558 & 0.916 & 0.733 & 1.314 \\
KIE & 0.006 & 0.117 & 0.276 & 0.422 & 0.502 & 0.638 & 0.720 & 0.729 & 0.746 & 0.789 & 0.783 & 0.818 & 0.545 & 0.993 & 0.825 & 1.512 \\
STU & 0.105 & 0.302 & 0.415 & 0.520 & 0.544 & 0.576 & 0.622 & 0.637 & 0.653 & 0.656 & 0.648 & 0.664 & 0.528 & 0.979 & 0.576 & 1.090 \\
\bottomrule
\end{tabular}
} %
\caption{Detailed performance and robustness results for the Fine-tuned Qwen-3B model on all 15 sub-tasks.}
\end{table*}

\subsection{Details of Res-Bench}

\subsubsection{Definitions of Core Capability Dimensions}
To gather a diverse and high-quality candidate pool, we began by aggregating samples from 13 established benchmarks. This included general-purpose MLLM evaluation suites  \cite{mmstar, seed}, as well as specialized benchmarks targeting specific skills such as visual perception \cite{blink, hrbench}, mathematical reasoning \cite{mathvista, mathverse}, chart understanding \cite{chartqapro}, spatial reasoning \cite{spatialeval}, Expert-level multimodal understanding \cite{mmmu, mmmupro}, robustness evaluation \cite{NaturalBench} ,and OCR \cite{OCRbench, ccocr}.

We now introduce the six core ability dimensions of our dataset.
\begin{itemize}
    \item \textbf{Coarse-grained Perception.} Coarse perception refers to the holistic comprehension of an image, focusing solely on its overarching theme or style without delving into specific content or fine-grained details. This dimension comprises two sub-capabilities: (1) image style; (2) image scene understanding.
    \item \textbf{Fine-grained Perception.} This dimension refers to localized image perception, which focuses exclusively on fine-grained details of an image without requiring holistic comprehension. The task demands highly precise recognition of local regions, necessitating clear cognitive processing of image fragments. This dimension comprises three sub-capabilities: (1) instance attribute; (2) instance location; (3) instance counting.
    \item \textbf{Instance Reasoning.} This dimension measures instance-level reasoning capability, including inter-instance relationship analysis and discriminative feature comparison. This dimension comprises two sub-capabilities: (1) cross-instance comparison; (2) spatial relation reasoning.
    \item \textbf{Logical Reasoning.} This dimension focuses on understanding structured information and processing abstract concepts, while deliberately excluding concrete entity recognition. This dimension comprises two sub-capabilities:
    (1) chart\&diagram reasoning; (2) science\&technology.
    \item \textbf{Mathmatics.} The dimension assesses visual mathematical reasoning, requiring the integration of visual inputs with symbolic or quantitative inference processes. This dimension comprises three sub-capabilities: (1) function caculation; (2) geometry; (3) stastical analysis.
    \item \textbf{OCR.} This dimension evaluates the model's OCR capabilities, encompassing both accurate text recognition in real-world scenarios (\eg, street-view scenarios, documents) and semantic comprehension of extracted content. This dimension comprises three sub-capabilities : (1) key information extraction; (2) scene-text understanding; (3) statistical analysis.
\end{itemize}

\subsubsection{More Examples of Res-Bench}

\begin{figure*}[htbp]
    \centering
  \includegraphics[width=\textwidth]{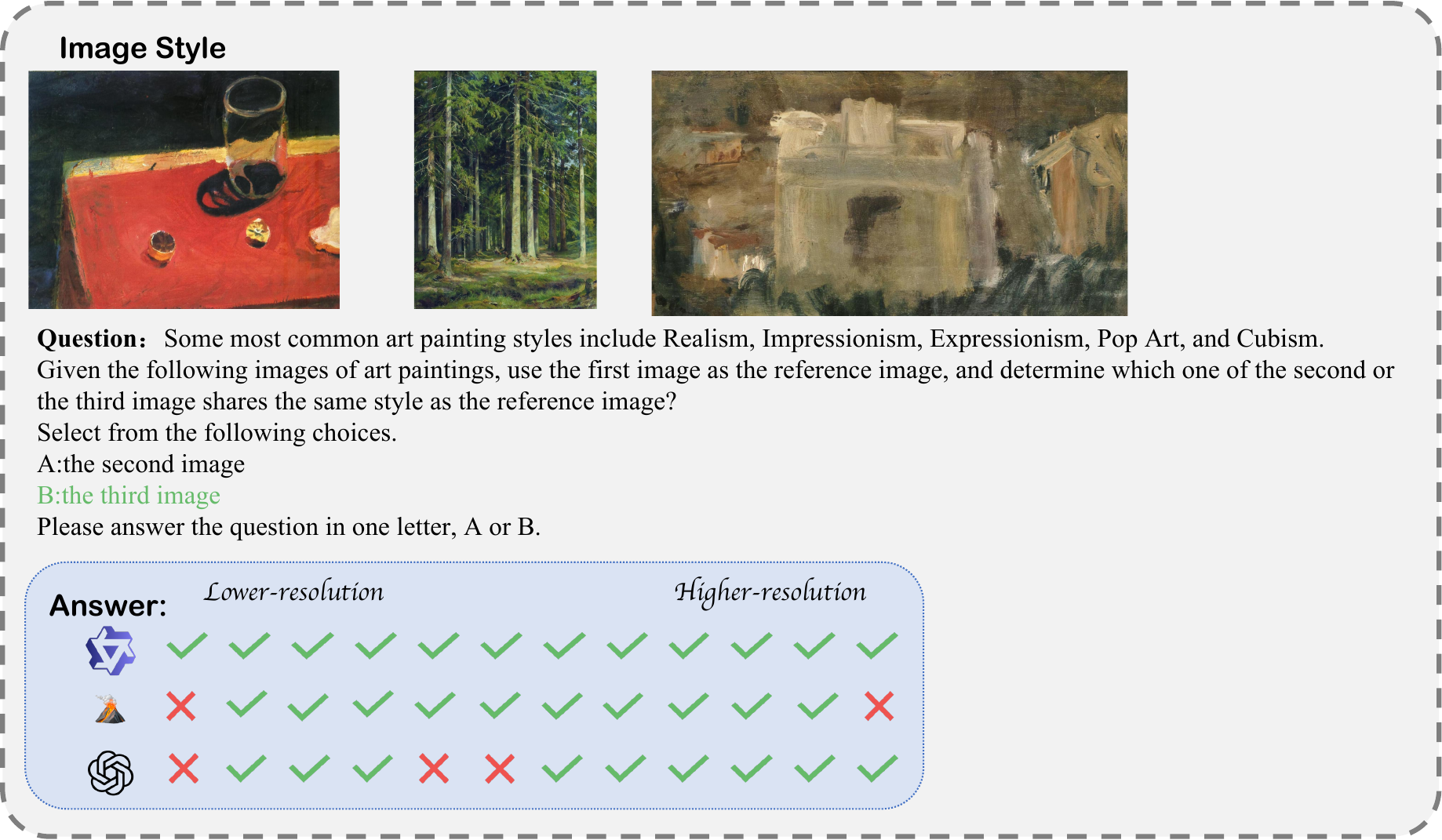}
      \caption {An example of the ``Image Style'' sub-task from the ``Coarse-grained Perception'' capability dimension.}
  \label{case1}
\end{figure*}

\begin{figure*}[htbp]
  \includegraphics[width=\textwidth]{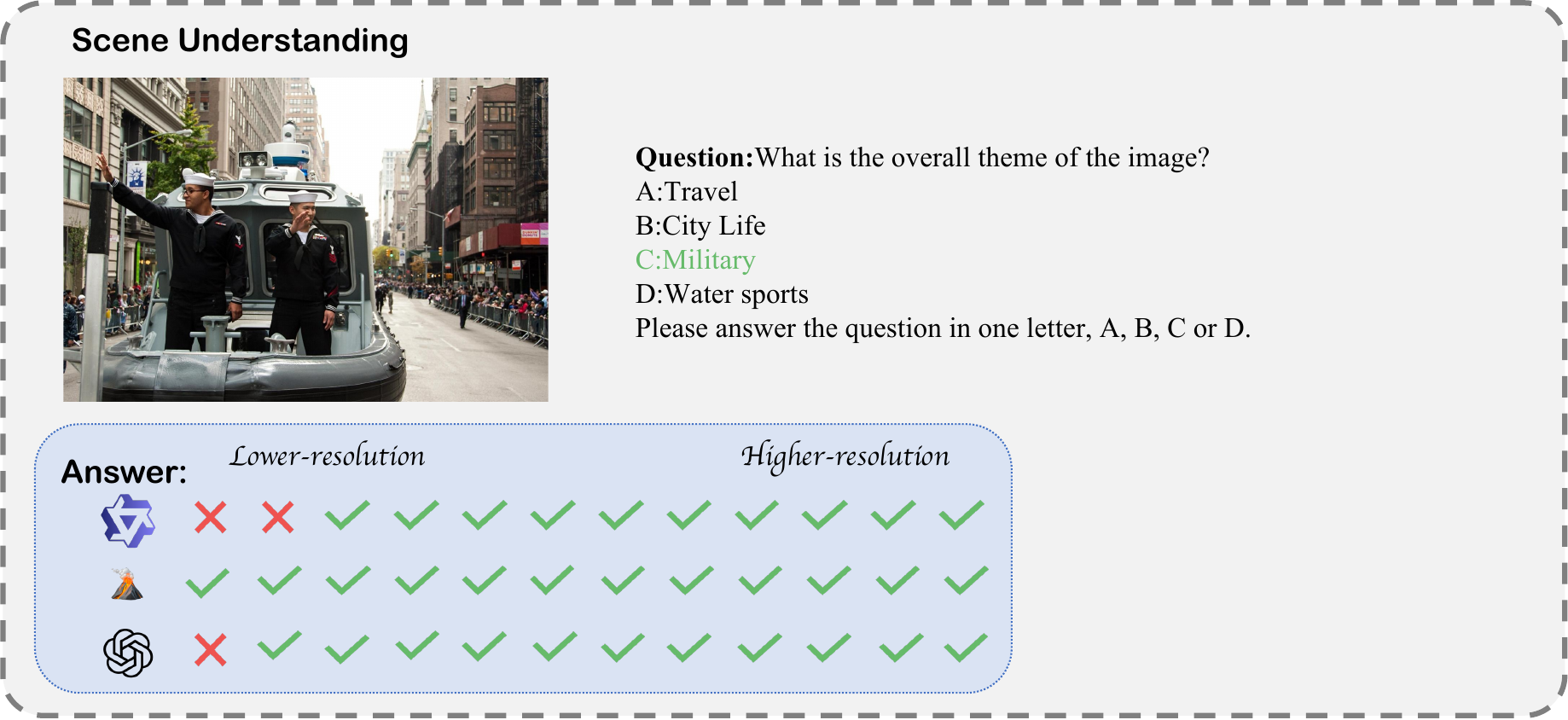}
      \caption {An example of the ``Scene Understanding'' sub-task from the ``Coarse-grained Perception'' capability dimension.}
  \label{case2}
\end{figure*}

\begin{figure*}[htbp]
  \includegraphics[width=\textwidth]{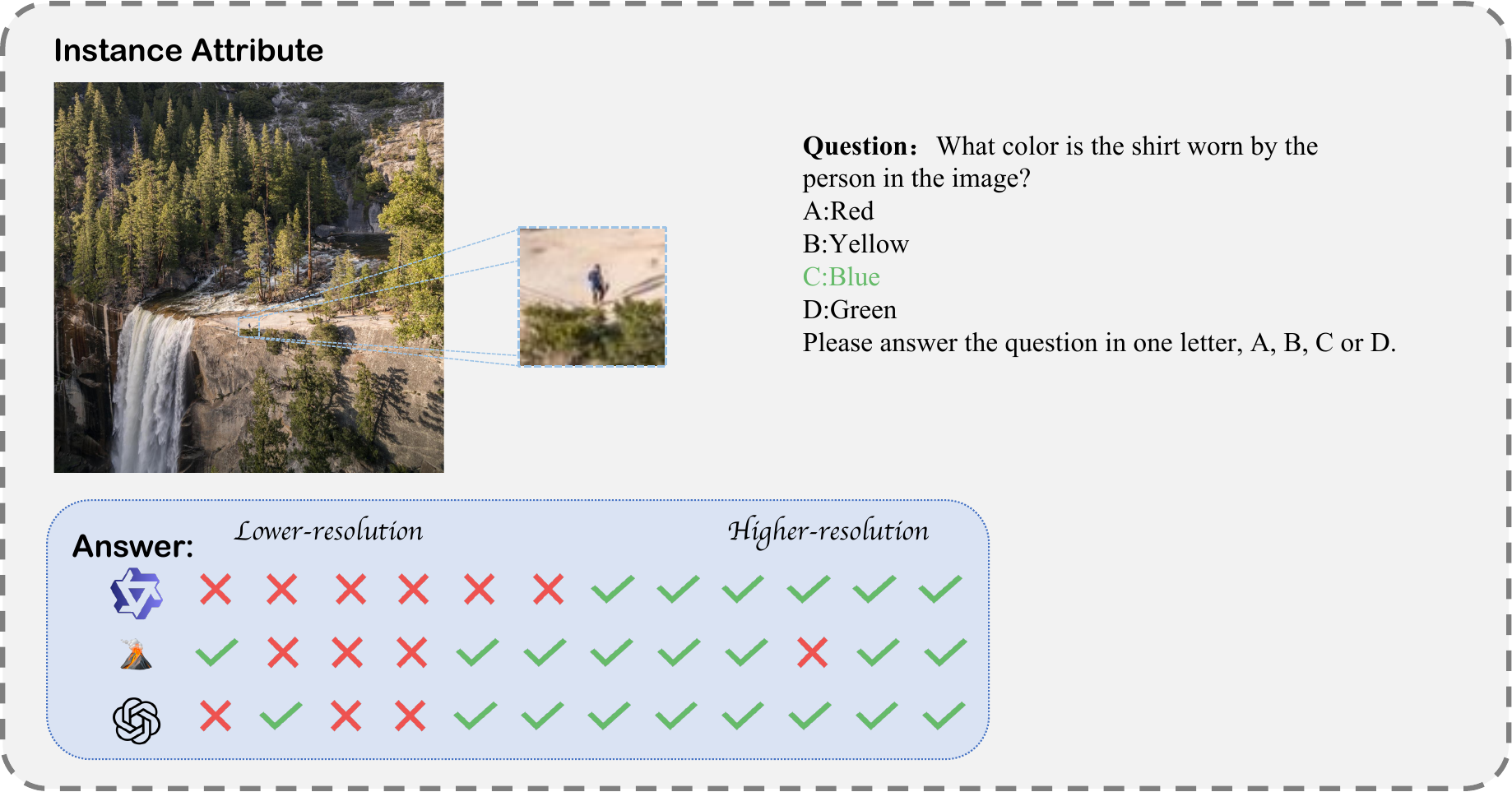}
      \caption {An example of the ``Instance Attribute'' sub-task from the ``Fine-grained Perception'' capability dimension.}
  \label{case3}
\end{figure*}

\begin{figure*}[htbp]
  \includegraphics[width=\textwidth]{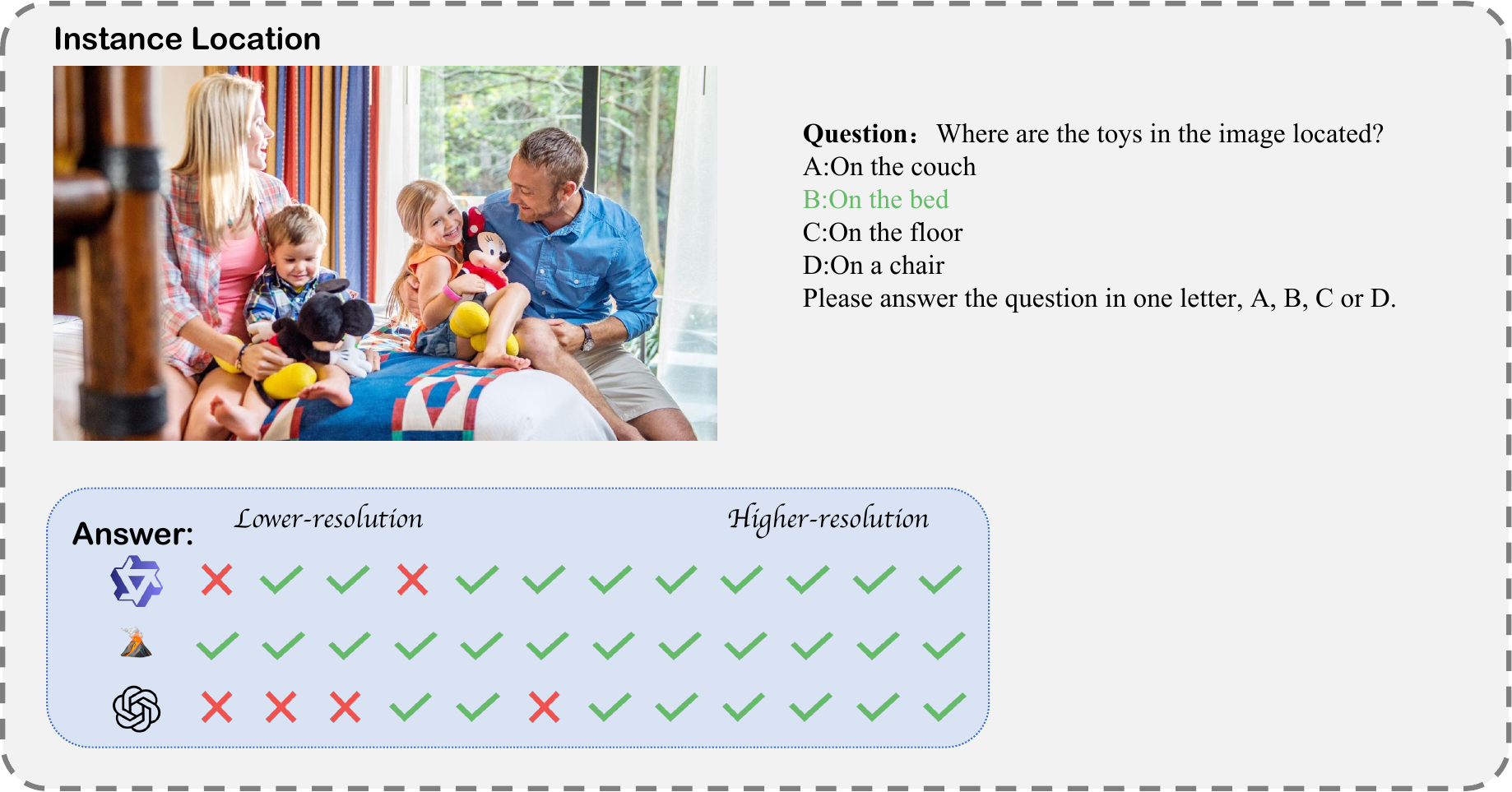}
      \caption {An example of the ``Instance Location'' sub-task from the ``Fine-grained Perception'' capability dimension.}
  \label{case4}
\end{figure*}

\begin{figure*}[htbp]
  \includegraphics[width=\textwidth]{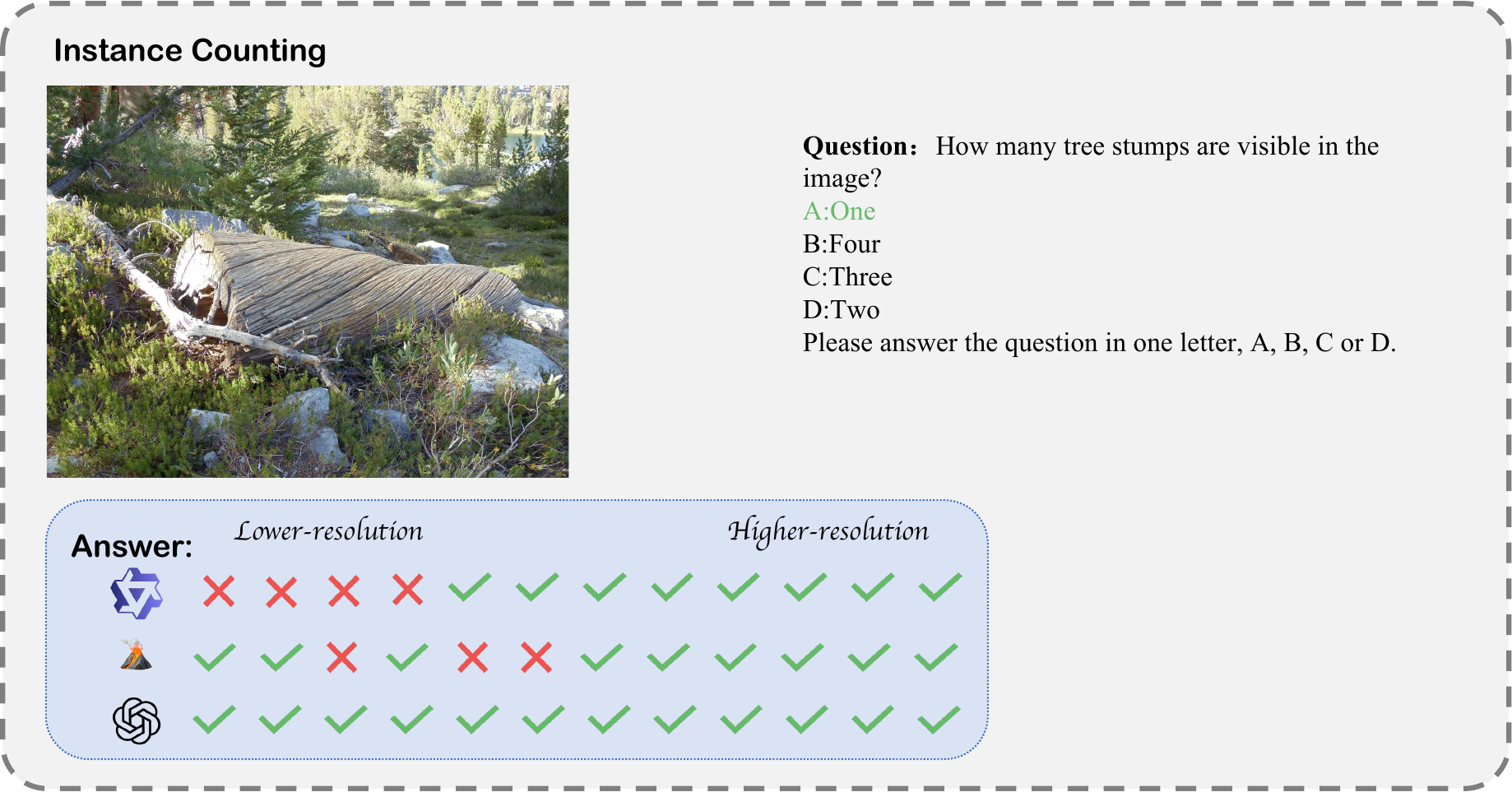}
      \caption {An example of the ``Instance Counting'' sub-task from the ``Fine-grained Perception'' capability dimension.}
  \label{case5}
\end{figure*}

\begin{figure*}[htbp]
  \includegraphics[width=\textwidth]{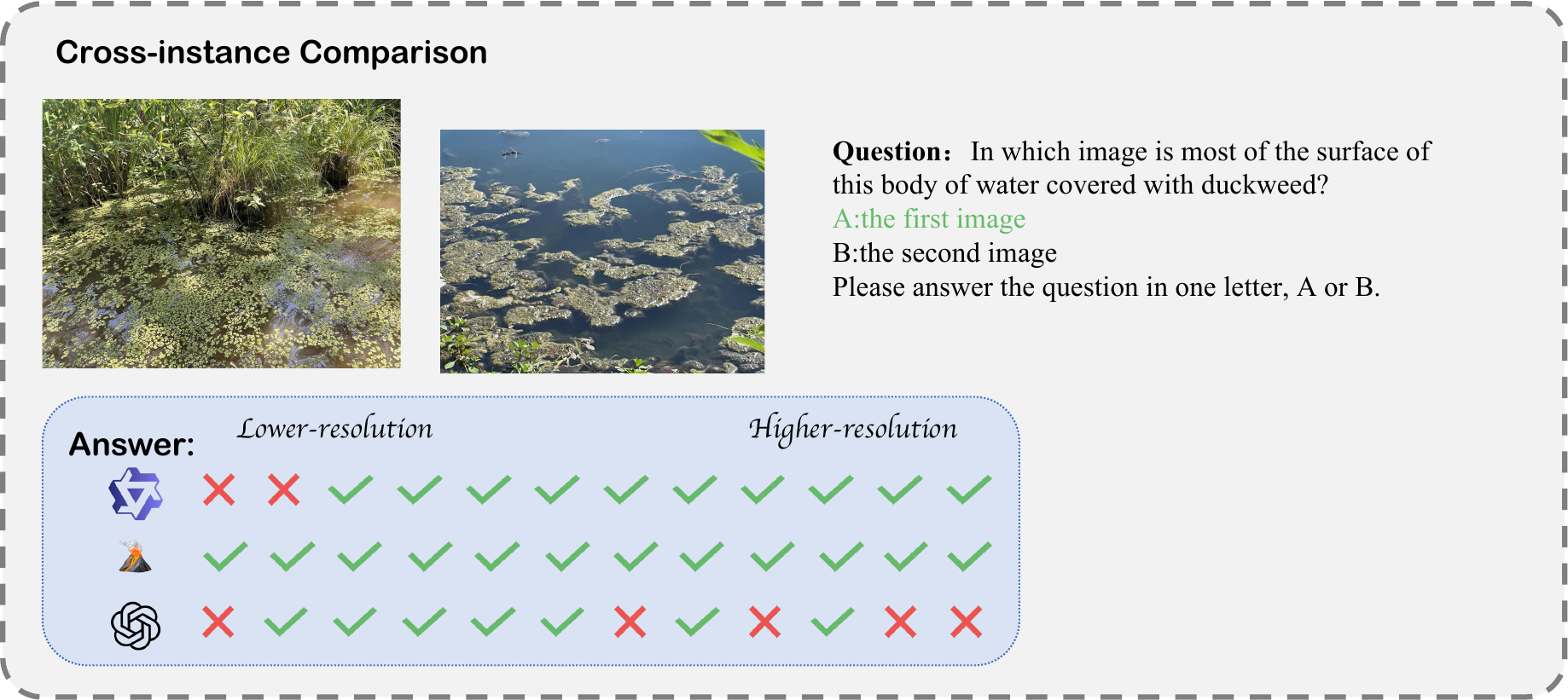}
      \caption {An example of the ``Cross-instance Comparison'' sub-task from the ``Instance Reasoning'' capability dimension.}
  \label{case6}
\end{figure*}

\begin{figure*}[htbp]
  \includegraphics[width=\textwidth]{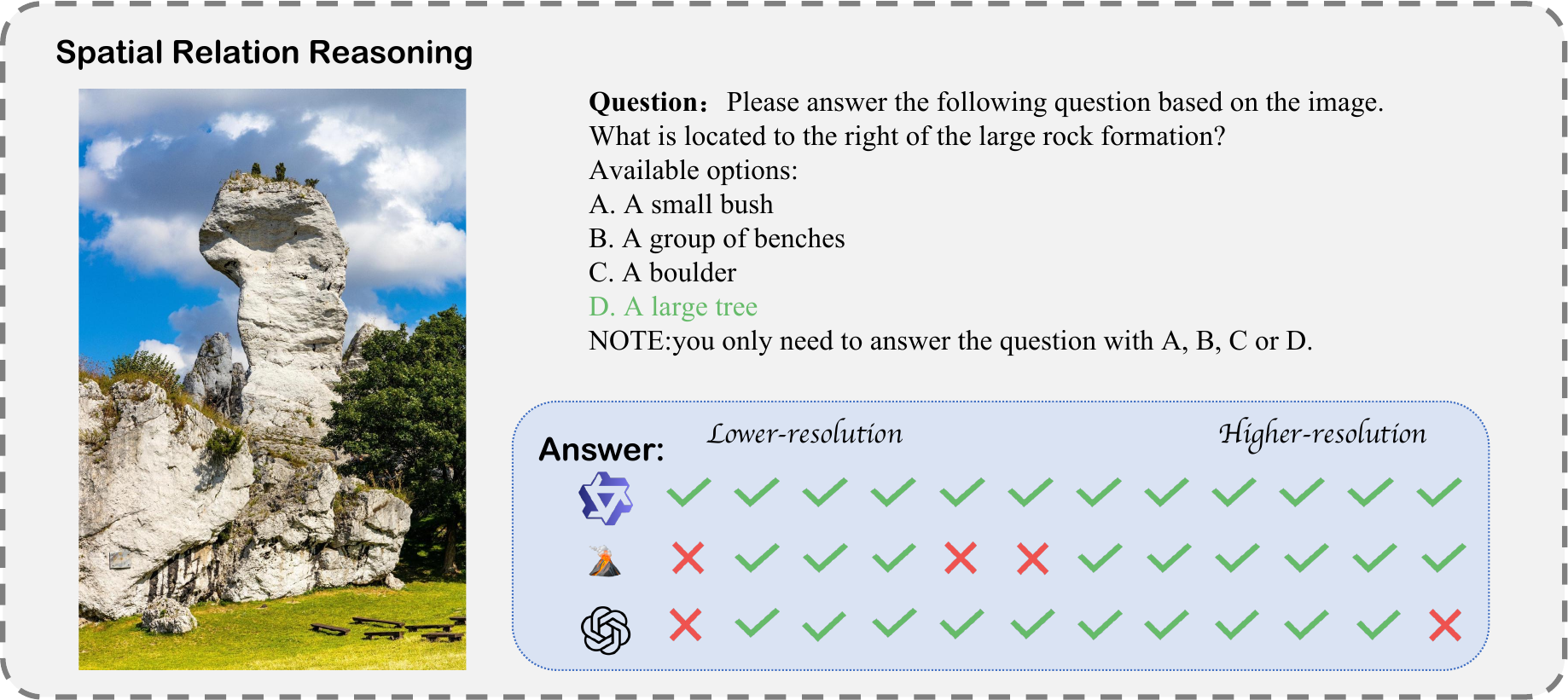}
      \caption {An example of the ``Spatial Relation Reasoning'' sub-task from the ``Instance Reasoning'' capability dimension.}
  \label{case7}
\end{figure*}

\begin{figure*}[htbp]
  \includegraphics[width=\textwidth]{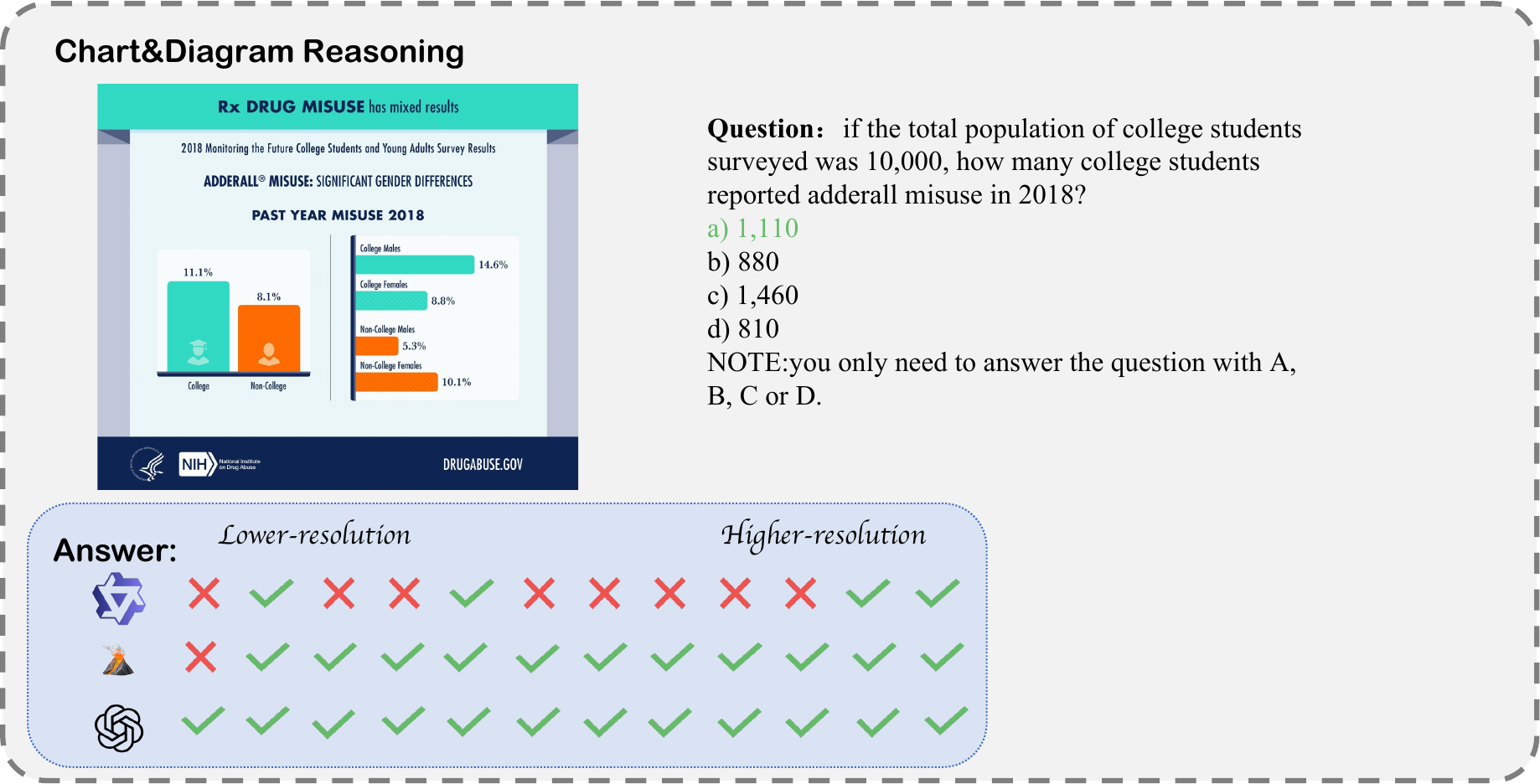}
      \caption {An example of the ``Chart\&Diagram Reasoning'' sub-task from the ``Logical Reasoning'' capability dimension.}
  \label{case8}
\end{figure*}

\begin{figure*}[htbp]
  \includegraphics[width=\textwidth]{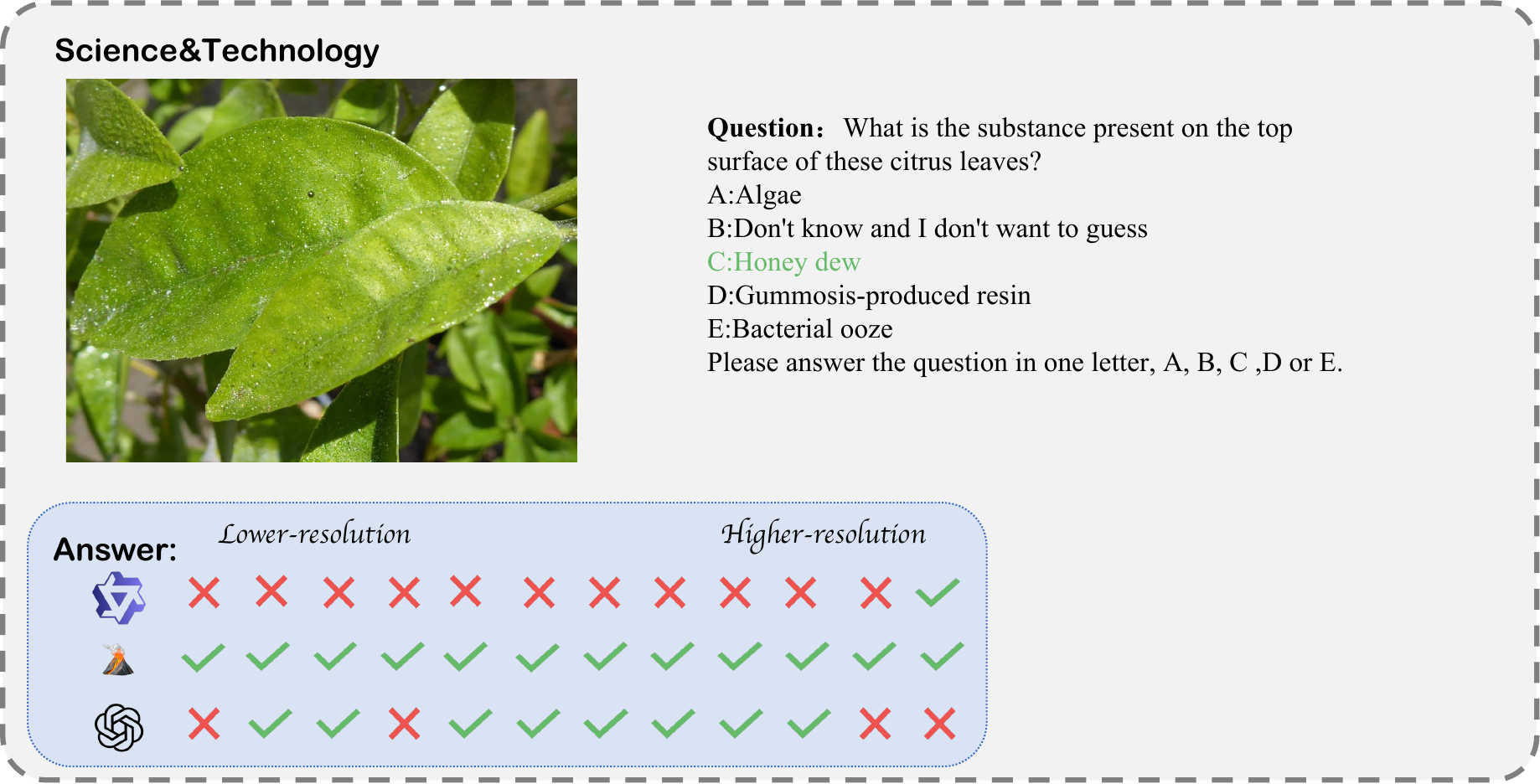}
      \caption {An example of the ``Science\&Technology'' sub-task from the ``Logical Reasoning'' capability dimension.}
  \label{case9}
\end{figure*}

\begin{figure*}[htbp]
  \includegraphics[width=\textwidth]{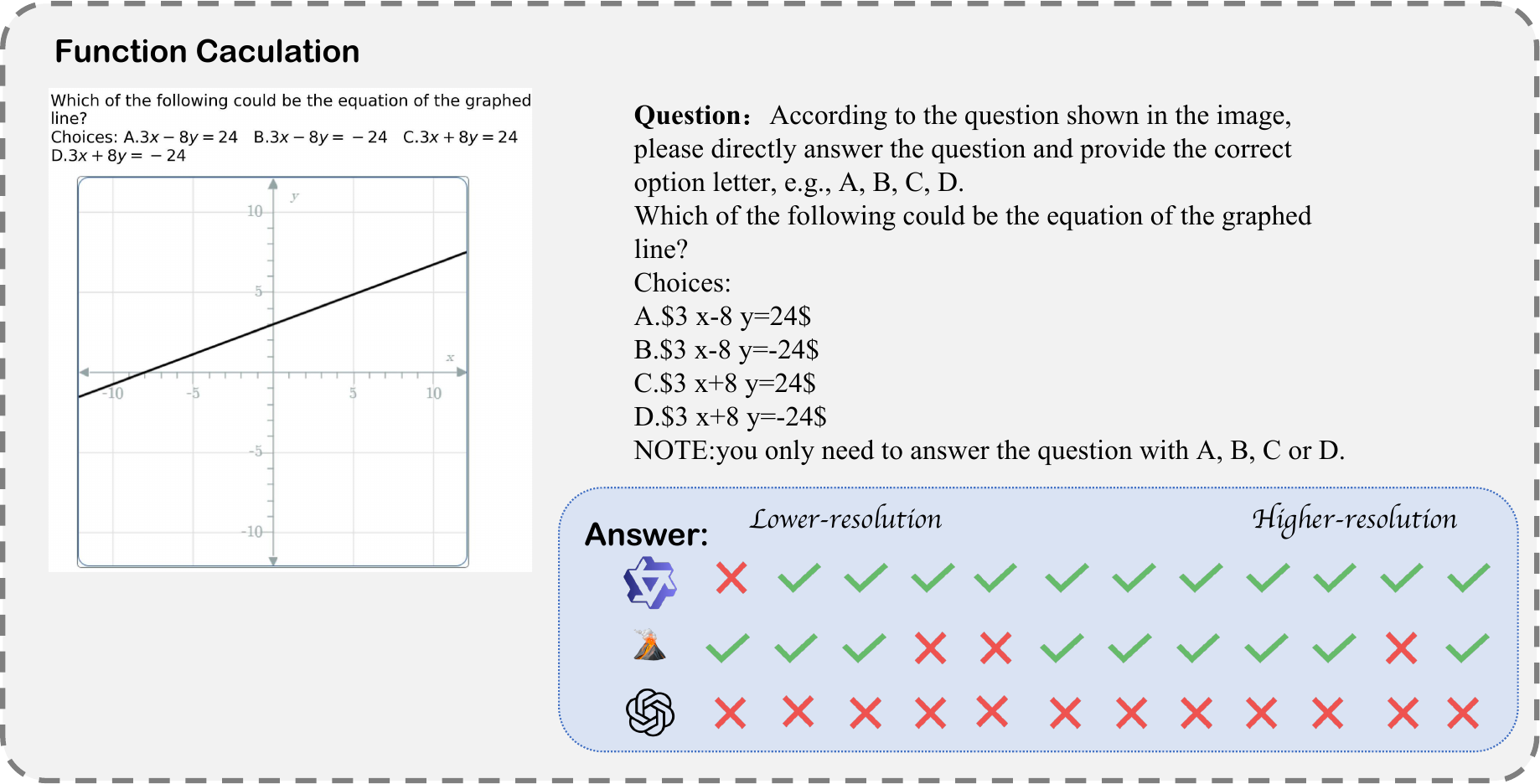}
      \caption {An example of the ``Function Caculation'' sub-task from the ``Mathematics'' capability dimension.}
  \label{case10}
\end{figure*}

\begin{figure*}[htbp]
  \includegraphics[width=\textwidth]{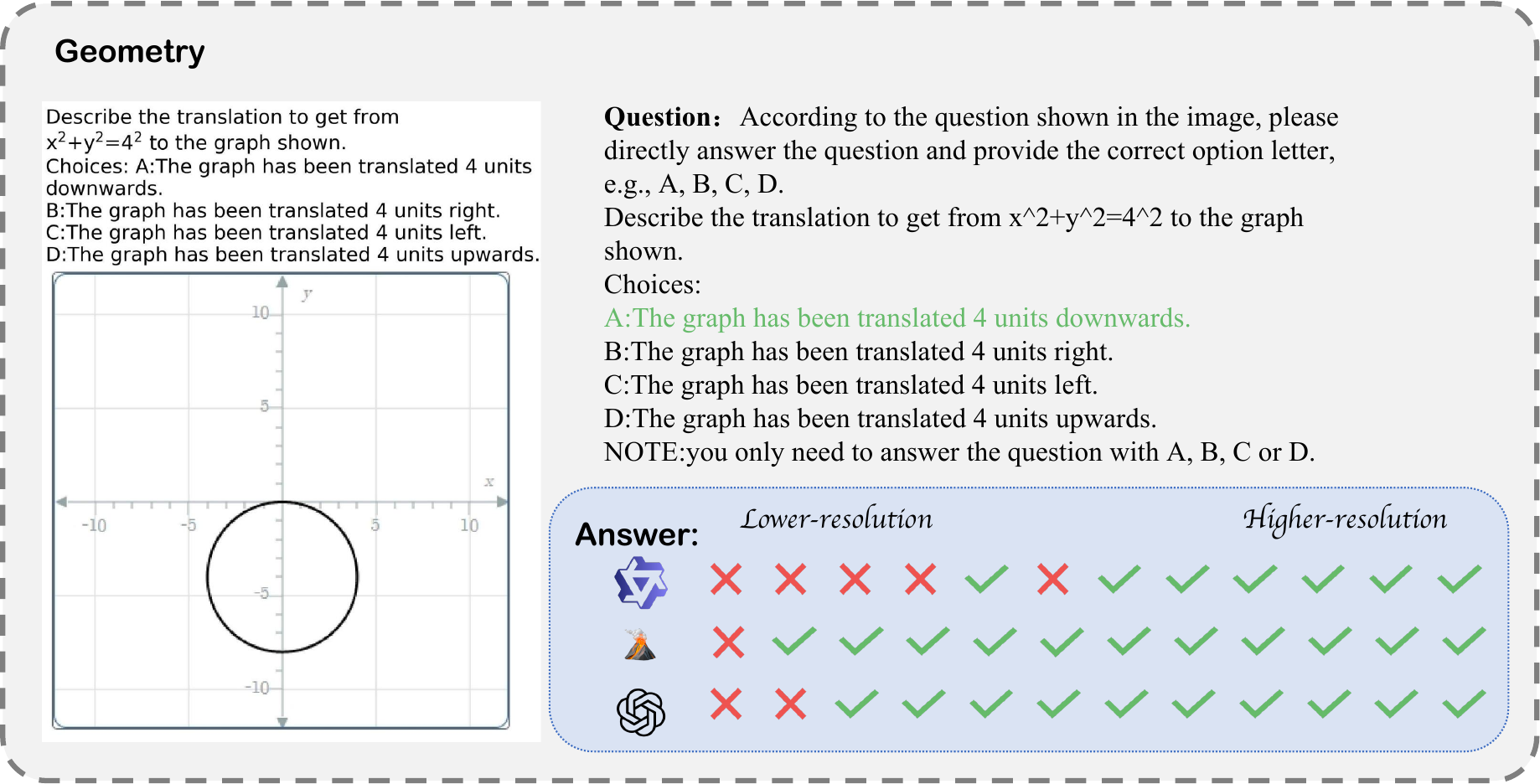}
      \caption {An example of the ``Geometry'' sub-task from the ``Mathematics'' capability dimension.}
  \label{case11}
\end{figure*}

\begin{figure*}[htbp]
  \includegraphics[width=\textwidth]{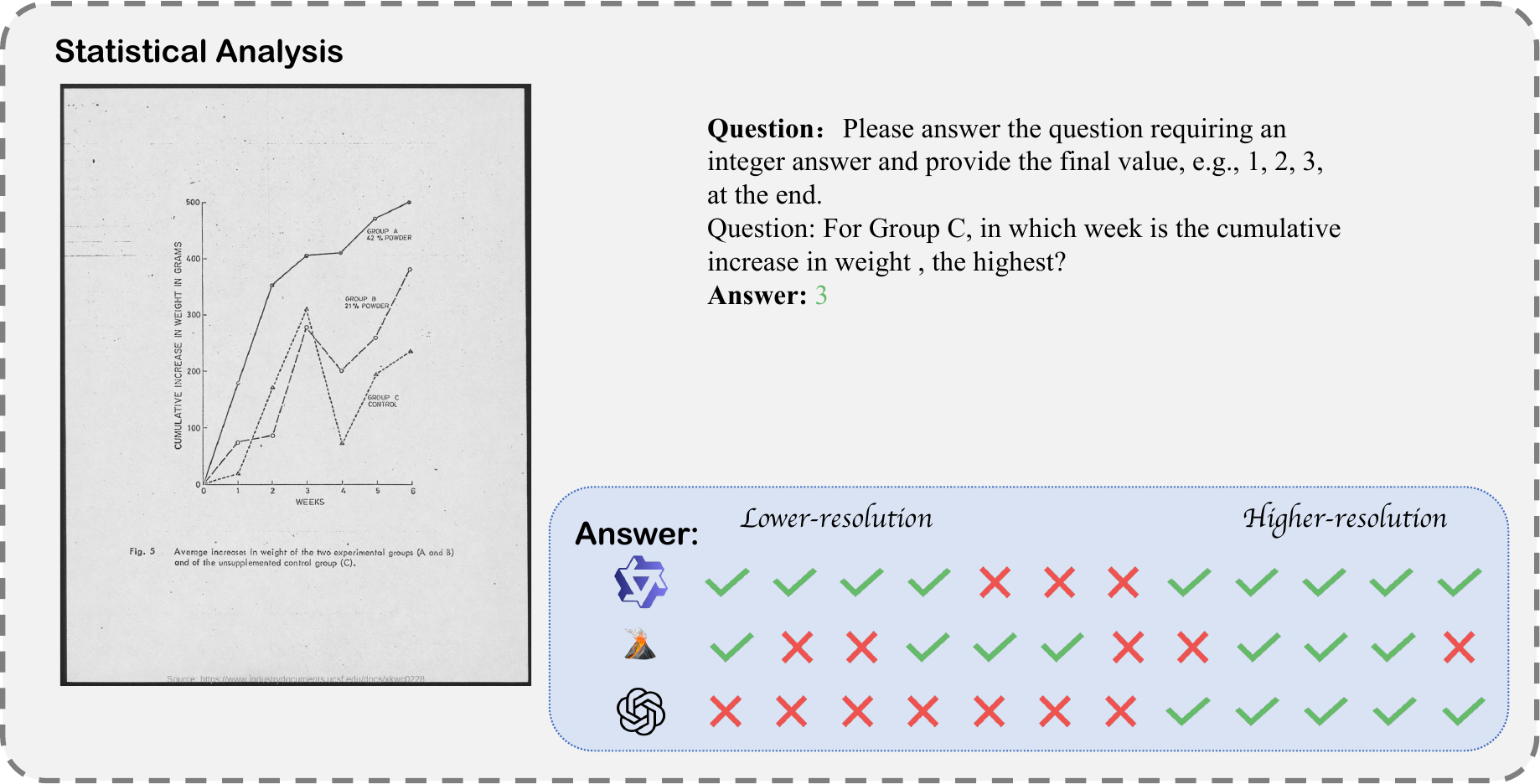}
      \caption {An example of the ``Statistical analysis'' sub-task from the ``Mathematics'' capability dimension.}
  \label{case12}
\end{figure*}

\begin{figure*}[htbp]
  \includegraphics[width=\textwidth]{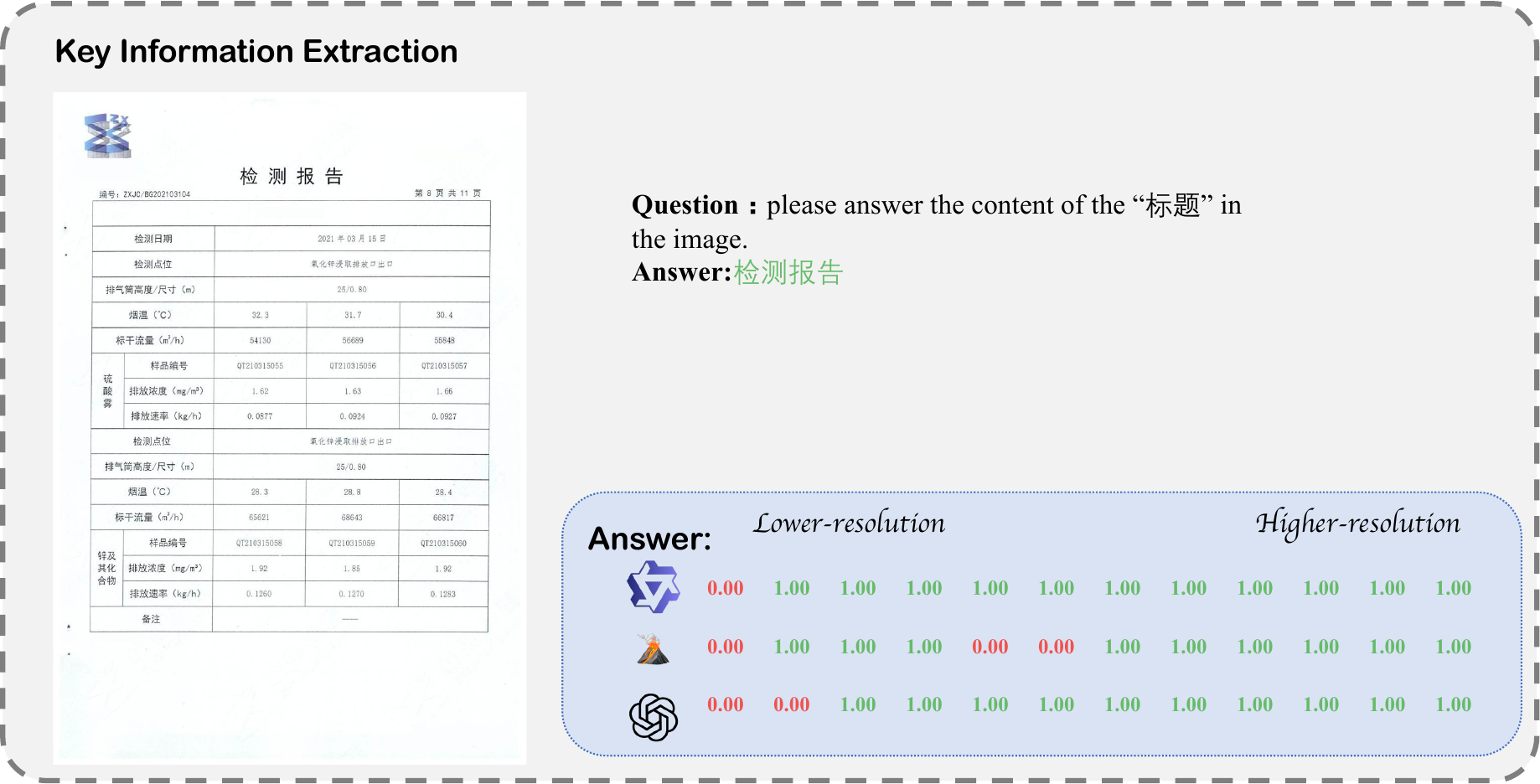}
      \caption {An example of the ``Key Information Extraction'' sub-task from the ``OCR'' capability dimension.}
  \label{case13}
\end{figure*}

\begin{figure*}[htbp]
  \includegraphics[width=\textwidth]{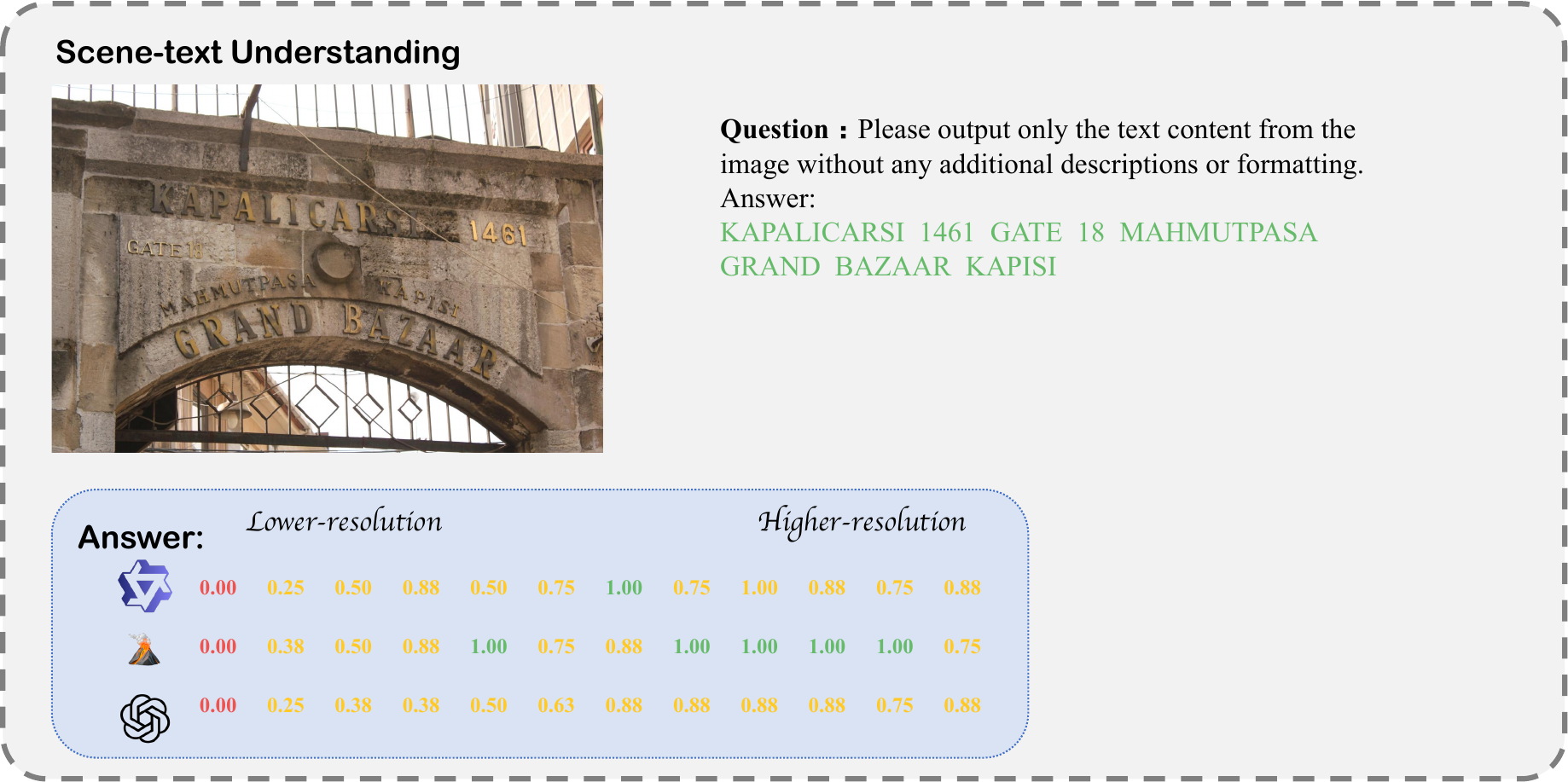}
      \caption {An example of the ``Scene-text Understanding'' sub-task from the ``OCR'' capability dimension.}
  \label{case14}
\end{figure*}

\begin{figure*}[htbp]
  \includegraphics[width=\textwidth]{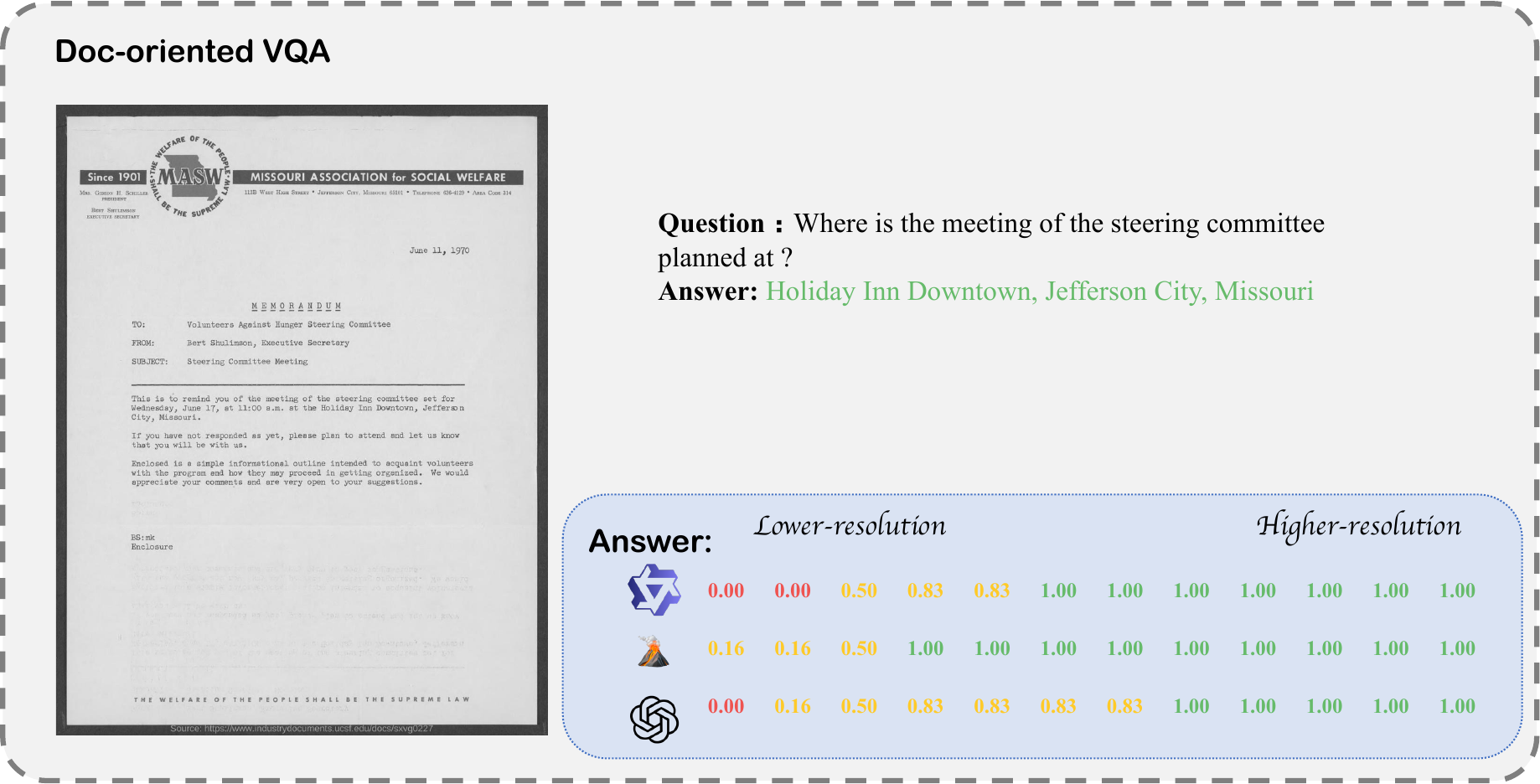}
      \caption {An example of the ``Doc-oriented VQA'' sub-task from the ``OCR'' capability dimension.}
  \label{case15}
\end{figure*}


\end{document}